\documentclass[11pt]{article}

\usepackage[final]{acl}
\usepackage{times}
\usepackage{latexsym}

\usepackage[T1]{fontenc}
\usepackage[utf8]{inputenc}

\usepackage{microtype}

\usepackage{inconsolata}

\usepackage{graphicx}

\usepackage{amsmath}
\usepackage{booktabs}
\usepackage{multirow}
\usepackage{makecell}
\usepackage{float}
\usepackage{arydshln}
\usepackage{cleveref}

\title{NOSE: Neural Olfactory-Semantic Embedding with Tri-Modal Orthogonal Contrastive Learning}

\author{
 \textbf{Yanyi Su\textsuperscript{1,2}\thanks{Work done during internship at DP Technology.}},
 \textbf{Hongshuai Wang\textsuperscript{2}},
 \textbf{Zhifeng Gao\textsuperscript{2}\thanks{Corresponding author.}},
 \textbf{Jun Cheng\textsuperscript{1,3,4}\footnotemark[2]},
\\
\textsuperscript{1}State Key Laboratory of Physical Chemistry of Solid Surface, \\ 
College of Chemistry and Chemical Engineering, Xiamen University, Xiamen, China \\
\textsuperscript{2}DP Technology, \\ 
\textsuperscript{3}Laboratory of AI for Electrochemistry (AI4EC), \\ 
Tan Kah Kee Innovation Laboratory (IKKEM), Xiamen, China \\
\textsuperscript{4}Institute of Artificial Intelligence, Xiamen University, Xiamen, China \\
\\
 \small{
 \textbf{Correspondence:}
   \href{mailto:gaozf@dp.tech}{gaozf@dp.tech}, 
  \href{mailto:chengjun@xmu.edu.cn}{chengjun@xmu.edu.cn}
 }
}

\begin{document}
\maketitle
\begin{abstract}
Olfaction lies at the intersection of chemical structure, neural encoding, and linguistic perception, yet existing representation methods fail to fully capture this pathway. Current approaches typically model only isolated segments of the olfactory pathway, overlooking the complete chain from molecule to receptors to linguistic descriptions. Such fragmentation yields learned embeddings that lack both biological grounding and semantic interpretability. We propose NOSE (Neural Olfactory-Semantic Embedding), a representation learning framework that aligns three modalities along the olfactory pathway: molecular structure, receptor sequence, and natural language description. Rather than simply fusing these signals, we decouple their contributions via orthogonal constraints, preserving the unique encoded information of each modality. To address the sparsity of olfactory language, we introduce a weak positive sample strategy to calibrate semantic similarity, preventing erroneous repulsion of similar odors in the feature space. Extensive experiments demonstrate that NOSE achieves state-of-the-art (SOTA) performance and excellent zero-shot generalization, confirming the strong alignment between its representation space and human olfactory intuition. Code and data are available at \url{https://github.com/Xianyusyy/NOSE}
\end{abstract}

\section{Introduction}
Among the major senses, olfaction is arguably the most challenging to digitize. Vision has pixels and hearing has spectra. These physical quantities maintain stable mappings to perception. But olfaction is different: the same molecule may activate different combinations of receptors, and human descriptions of odors are highly subjective. Olfaction initiates with molecular-receptor binding, propagates through neural signal transduction, and culminates in perceptual formation within the brain \citep{buck1991novel,su2009olfactory,sobel1998sniffing,lapid2011neural}.   

To provide context for a broader audience, we briefly introduce the key concepts. In molecular informatics, SMILES (Simplified Molecular-Input Line-Entry System) \citep{weininger1988smiles} is a standard notation that encodes molecular graphs as ASCII strings; for example, \texttt{CCO} represents ethanol and \texttt{c1ccccc1} represents benzene. Olfactory receptors (ORs) are G-protein-coupled receptor proteins located in the nasal epithelium. The human genome encodes approximately 400 functional ORs \citep{malnic1999combinatorial,buck2004olfactory}, each responding to specific molecular features, with sequences typically spanning $\sim$310 amino acids. When an odorant molecule binds to a receptor, it triggers a neural signal cascade that ultimately produces a conscious percept. The central computational challenge in this domain is therefore to predict, given a molecular structure, the perceptual attributes that humans would report, ranging from basic properties such as detection threshold, intensity, and pleasantness, to high-level semantic descriptors such as ``floral'' or ``sweet.'' Representative inputs for each modality are illustrated in Table~\ref{tab:input_examples}.

\begin{table}[t]
  \centering
  \small
  \begin{tabular}{p{1.5cm}p{5.5cm}}
    \toprule
    \textbf{Modality} & \textbf{Input Example} \\
    \midrule
    Molecule & \texttt{CC1CCCCCCCCCCCCC(=O)C1} (Muscone) \\
    \addlinespace[2pt]
    Receptor & \texttt{MRENNQSSTLEFILLGVTG...} (OR5A2, 312 amino acids) \\
    \addlinespace[2pt]
    Description & ``musk; powdery; sweet; floral'' \\
    \bottomrule
  \end{tabular}
  \caption{Representative input examples for the three pre-training modalities. Molecule and receptor inputs correspond to Muscone and its cognate receptor OR5A2, respectively.}
  \label{tab:input_examples}
\end{table}

However, existing methods typically model only fragments of this pathway \citep{jiang2025dual,lee2023principal,chithrananda2024mapping,gupta2021odorify}. They focus either solely on molecular structure or learn only molecule-description/receptor correspondences, but have never captured the complete chain from molecule to receptor to semantics within a unified framework. A more fundamental issue lies in the task formulation: mainstream methods treat odor prediction as a classification problem, predicting whether a given molecule belongs to "floral" or "fruity". This discretization leads to two consequences. First, it destroys the continuity of odor space. "minty" and "cooling" are highly correlated in human perception, but under the classification framework they are merely two independent labels, leaving the model unable to learn such associations. Similarly, sequence-similar receptors often mediate related odor responses, yet this continuity is also ignored. Second, classification objectives erode the molecular representation itself. When models are forced to fit classification boundaries for odor labels, they often discard information that is useless for classification but crucial for molecular structure. This causes models to perform reasonably on known odor categories but fail to generalize to novel molecules or descriptions.   

These observations motivate us to rethink how olfactory representations should be constructed. We propose NOSE (Neural Olfactory-Semantic Embedding), a tri-modal learning framework designed with molecules as the central hub. The core insight stems from a pragmatic observation: although "molecule-receptor-odor description" triplets are extremely scarce, "molecule-receptor" and "molecule-description" bimodal data can be obtained separately. Molecules are the sole intersection of both, serving as a hub to bridge receptor information and odor semantic information into a unified representation space. A question naturally arises: if receptor features and odor semantic features are simultaneously injected into molecular representations, will the three modalities interfere with and overwrite each other? Our solution is orthogonal injection, forcing the two types of features to occupy orthogonal subspaces in the representation space. Receptor information and odor semantic information are superimposed onto molecular representations as mutually independent increments, preserving the integrity of molecular structure while achieving implicit tri-modal alignment. Another practical challenge is the sparsity of odor description data: each molecule is typically annotated with only a few descriptors, while semantically similar words are often treated as irrelevant labels. To address this, we leverage large language models to mine odor semantic proximity relationships among descriptors, expanding isolated labels into continuous odor semantic neighborhoods and mitigating the false negative problem in contrastive learning.  

Our contributions span three aspects:  

\textbf{(1) Data Infrastructure}: We integrate and curate multi-source odor description and receptor data, constructing the first large-scale pre-training dataset supporting tri-modal learning along with accompanying evaluation benchmarks.  

\textbf{(2) Representation Learning Framework}: We propose an orthogonal feature injection mechanism that achieves global alignment and decoupled representations of molecules, receptors, and semantics without relying on triplet annotations.   

\textbf{(3) Semantic Topology Preservation}: We design a LLM-enhanced contrastive learning strategy that transforms the discrete odor label space into a continuous odor semantic manifold, mitigating false negative issues caused by label sparsity.  

\begin{figure*}[t]
  \includegraphics[width=\textwidth]{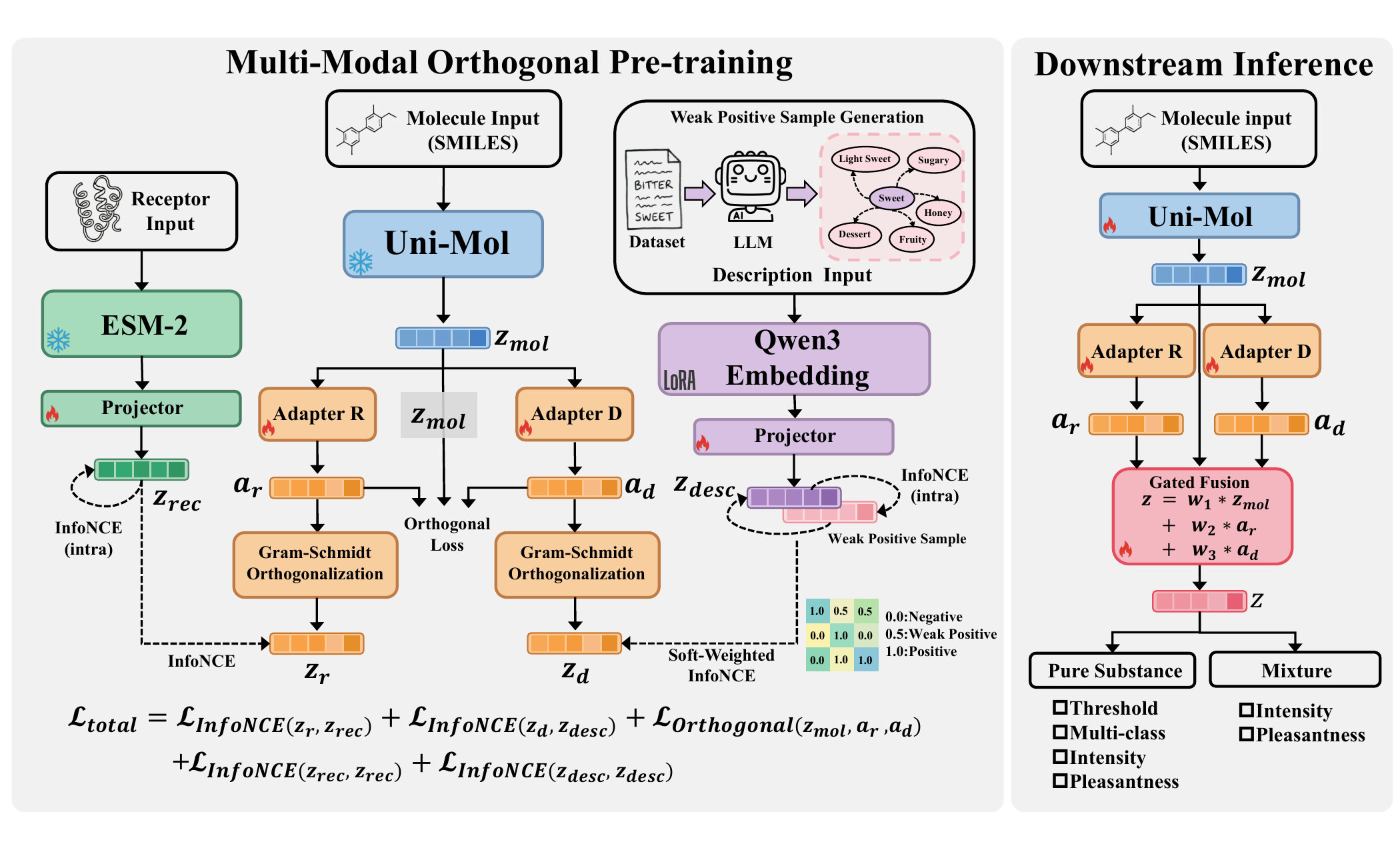}
  \caption{Subscripts $r$ and $d$ denote receptor and description modalities, respectively. \textbf{(Left)} Multimodal Orthogonal Pre-training: Molecular representations $z_{mol}$ are extracted by a frozen Uni-Mol encoder, receptor embeddings $z_{rec}$ are obtained from ESM-2 with a trainable projection layer, and odor semantic descriptions $z_{desc} $ are extracted by LoRA-finetuned Qwen3 Embedding after LLM-based weak positive augmentation. The molecular embedding is decomposed via dual adapters into a receptor-aligned component ($a_r$) and a description-aligned component ($a_d$), which are then orthogonalized through Gram-Schmidt to yield $z_r$ and $z_d$. Training objectives include: receptor-molecule InfoNCE loss, description-molecule soft-weighted InfoNCE loss (with positive/weak-positive/negative sample weights of 1.0/0.5/0.0), intra-modal InfoNCE loss, and orthogonality constraints among $z_{mol}$, $a_r$, and $a_d$. \textbf{(Right)} Downstream Inference: Inference requires only the molecular encoder and adapters, with the final representation $Z = w_1 \cdot z_{mol} + w_2 \cdot a_r + w_3 \cdot a_d$, supporting both pure substance and mixture perception tasks.}
  \label{main_image}
\end{figure*}

\section{Related Work}
Predicting odor perception is extremely challenging because it relies on multiple interacting information sources. Early work focused on quantitative structure-activity relationships (QSAR), attempting to predict odor perception solely from molecular structure \citep{jiang2025dual,sharma2025navigating,taleb2024can,shin2023optimizing,ravia2020measure,tran2019deepnose,keller2017predicting}. However, structure-odor relationships are inherently nonlinear \citep{sell2006unpredictability}: minor structural changes can cause dramatic perceptual shifts, while structurally dissimilar molecules may smell alike.   

To address this, recent research has begun incorporating auxiliary modalities (odor semantics \citep{tom2025does,lee2023principal}, receptor information \citep{mcconachie2025low,wakutsu2025molecular,chithrananda2024mapping,gupta2021odorify}). However, these methods rely on strongly supervised paradigms, reducing representation learning to multi-label classification or binary receptor matching. This forces models to learn classification boundaries rather than continuous representations, causing molecular, receptor, and semantic features to become entangled in the latent space. Beyond these limitations, even with auxiliary modalities, these methods do not cover the complete olfactory perception pathway from molecules through receptors to semantic descriptions. 

Another limitation of existing methods is the neglect of intra-modal topological structure. On the semantic side, existing methods treat odor descriptors as discrete labels, ignoring the continuous semantic relationships among descriptors. For instance, "lemon" and "sour" should be adjacent in odor space. On the receptor side, binary binding prediction ignores the evolutionary homology and family hierarchy among receptor sequences. Such discretization discards intra-modal relationships as core priors, leaving models to learn only isolated mapping rules and unable to construct continuous olfactory spaces consistent with semantic intuition and biological mechanisms.

\begin{table*}[t]
  \centering
  \small
  \begin{tabular}{llll}
    \toprule
    \textbf{Category} & \textbf{Task} & \textbf{Data Format} & \textbf{Source (Size)} \\
    \midrule
    \multirow{3}{*}{\shortstack[l]{Pretraining\\Corpora}} 
      &Contrastive & SMILES–Receptor Sequence pairs & \citeauthor{lalis2024m2or}, etc. (3,877) \\
      \cmidrule(lr){2-4}
      & \multirow{2}{*}{Contrastive} & SMILES–Odor Description pairs & \citeauthor{arctander2017perfume}, etc. (88,512) \\
      \cmidrule(lr){3-4}
      &  & SMILES–Odor Description pairs & LLM-augmented\textsuperscript{*} (2,567,558) \\
    \midrule
    \multirow{6}{*}{\shortstack[l]{Basic\\Perception}} 
      & Threshold & SMILES $\to$ $[-2.78, 6.11]$ & \citeauthor{abraham2012algorithm} (268) \\
      \cmidrule(lr){2-4}
      & \multirow{2}{*}{Pleasantness} & SMILES $\to$ $[0, 100]$ & \citeauthor{keller2016olfactory} (474) \\
      \cmidrule(lr){3-4}
      &  & SMILES $\to$ $[-1, 1]$ & \citeauthor{sagar2023high} (160) \\
      \cmidrule(lr){2-4}
      & \multirow{3}{*}{Intensity} & SMILES $\to$ $[0, 100]$ & \citeauthor{ravia2020measure} (248) \\
      \cmidrule(lr){3-4}
      &  & SMILES $\to$ $[0, 100]$ & \citeauthor{keller2016olfactory} (474) \\
      \cmidrule(lr){3-4}
      &  & SMILES $\to$ $[-1, 1]$ & \citeauthor{sagar2023high} (160) \\
    \midrule
    \multirow{3}{*}{\shortstack[l]{Semantic\\Description}} 
      & Descriptor Classification & SMILES $\to$ 138 classes & \citeauthor{tom2025does} (4,814) \\
      \cmidrule(lr){2-4}
      & \multirow{2}{*}{Descriptor Strength} & SMILES $\to$ $[0, 100]$ (21-dim) & \citeauthor{keller2016olfactory} (474) \\
      \cmidrule(lr){3-4}
      &  & SMILES $\to$ $[-1, 1]$ (15-dim) & \citeauthor{sagar2023high} (160) \\
    \midrule
    \multirow{2}{*}{\shortstack[l]{Mixture\\Perception}} 
      & Mixture Intensity & Mixture $\to$ $[0, 10]$ & \citeauthor{ma2021dataset} (6,660) \\
      \cmidrule(lr){2-4}
      & Mixture Pleasantness & Mixture $\to$ $[0, 10]$ & \citeauthor{ma2021dataset} (6,660) \\
    \bottomrule
  \end{tabular}
  \caption{Overview of pretraining corpora and downstream evaluation tasks. 
Descriptor Classification is a multi-label classification task, while 
Descriptor Strength involves multi-dimensional regression. $^*$We leverage an LLM to scale up the pretraining corpora by orders of magnitude.}
  \label{tab:datasets}
\end{table*}

\section{Methodology}
\subsection{Dataset}
\subsubsection{Downstream task dataset}
Olfactory perception is a multi-level cognitive process. To comprehensively evaluate the quality of olfactory representations, we design an evaluation benchmark spanning three levels: \textbf{(1) Basic Perceptual Attribute Prediction}: The most fundamental human perception of odors include "can it be smelled" (threshold), "how strong is it" (intensity), and "is it pleasant" (pleasantness). \textbf{(2) Semantic Description Prediction}: This level examines whether the model can capture the mapping between molecular structure and high-level semantic concepts such as "creamy" and "grassy." \textbf{(3) Mixture Perception Prediction}: Targeting the ubiquitous mixed odors in real-world scenarios, this level evaluates the model's capability to capture complex nonlinear intermolecular interactions such as masking and synergy. Based on this framework, we integrate six public datasets: \citet{abraham2012algorithm}, GS-LF \citep{lee2023principal,sanchez2019machine,tom2025does}, \citet{ravia2020measure}, \citet{keller2016olfactory}, \citet{sagar2023high}, and \citet{ma2021dataset}, constructing 11 downstream tasks (see Table~\ref{tab:datasets} and Section~\ref{Downstream} for details).

\subsubsection{Pre-training dataset}

\textbf{\textit{Receptor Data}}: We integrate multi-source olfactory receptor data including Pred-O3 \citep{ollitrault2024pred}, OlfactionBase \citep{sharma2022olfactionbase}, M2OR \citep{lalis2024m2or}, as well as literature-derived data from \citet{mainland2015human} and \citet{ahmed2018molecular}. We construct an olfactory receptor-ligand interaction dataset encompassing extensive biological specificity. Data cleaning and standardization procedures are conducted. Non-human source data and entries that cannot be matched to standard sequences are rigorously excluded. All receptor IDs are converted to protein sequences. For entries with failed queries, manual tracing and sequence completion are performed. After removing duplicate entries across datasets, we construct a SMILES-receptor dataset containing 3,877 activation pairs.

\textbf{\textit{Description Data}}: We integrate 11 mainstream data sources spanning public databases, academic literature, and industrial catalogs to construct a large-scale comprehensive semantic dataset. Specifically, this encompasses public resources including \citet{arctander2017perfume}, \citet{GoodScentsCompany}, OlfactionBase \citep{sharma2022olfactionbase}, \citet{sanchez2019machine}, Flavornet \citep{acree2004flavornet}, FlavorDB \citep{garg2018flavordb}, AromaDb \citep{kumar2018aromadb}, deep learning benchmark datasets \citep{sharma2021smiles}, as well as industrial catalogs such as \citet{IFRAGlossary} and \citet{SigmaAldrich}. Additionally, we introduce odorless molecules from \citet{mayhew2022transport} as negative sample supplements.   

The total volume of raw data amounts to approximately 140,000 entries. We implement a rigorous cleaning pipeline, including removal of invalid molecular structures, unification of ambiguous semantic labels, and correction of distributional biases in specific categories. This process yields 88,512 valid data pairs. We retain differentiated descriptions from multiple sources for the same molecule to reflect the inherent subjective diversity of odor perception. To address the challenge of descriptor sparsity, we leverage the DeepSeek \citep{guo2025deepseek} model to construct an odor semantic space and mine weak positive samples for all the 1,086 odor descriptors. The final dataset encompasses 9,513 unique molecules, and after semantic augmentation, the total scale is significantly expanded to 2,567,558 SMILES-olfactory descriptor pairs. Details of data cleaning, SMILES standardization, and molecular overlap analysis are provided in Appendix~\ref{appendix:data_processing}.

\subsection{Encoder Selection}
\textbf{Molecular Encoder}: To capture the spatial conformations critical for receptor binding, we employ Uni-Mol \citep{zhou2023uni} as the molecular encoder. Pre-trained on 209 million molecular conformations in 3D, this model is capable of modeling stereochemical geometric information underlying molecule-receptor interactions. 
\textbf{Receptor Encoder}: We utilize the ESM-2(650M) \citep{lin2023evolutionary} to extract receptor sequence features. Pre-trained on large-scale protein databases, this model can accurately capture the implicit structural patterns from one-dimensional amino acid sequences. 
\textbf{Odor Descriptor Encoder}: To capture the rich semantics of odor descriptors, we adopt the Qwen3 Embedding \citep{zhang2025qwen3} model (8B). This model possesses strong generalization capabilities for general text and provides robust representations for subsequent injection of olfactory domain knowledge.
\subsection{The NOSE Framework}
Given that Uni-Mol and ESM-2 possess native representation capabilities for molecular and receptor features, we freeze their parameters to avoid disrupting their high-quality domain distributions. To address the misalignment between generic semantics and olfactory space distribution in Qwen3 Embedding \citep{kurfali2025representations,zhong2024sniff} (see section~\ref{Olfactory_Concepts} for details), we employ LoRA \citep{hu2022lora} for adaptive fine-tuning. This guides the model's representation manifold to migrate from generic contexts to the olfactory perception domain, thereby achieving precise cross-modal semantic alignment. 
\subsubsection{Projection Heads and Deep Adapters}
For ESM-2 and Qwen3 Embedding, we employ standard nonlinear projection heads. This module follows the design of SimCLR \citep{chen2020simple}. For Uni-Mol, we design a deep projection adapter based on ResMLP \citep{touvron2022resmlp}, adopting the Pre-LN \citep{xiong2020layer} structure. Furthermore, considering the significant disparity in data scale between protein sequences and odor descriptions, we adopt differentiated configurations for adapters of these two modalities. The description adapter employs a high-capacity 12-layer inverted bottleneck structure to fit the rich textual data, whereas the receptor adapter introduces a bottleneck structure with high dropout rates to prevent overfitting on sparse receptor data.
\subsubsection{Orthogonal Mechanism}
To prevent feature redundancy during multi-modal contrastive learning, we employ a combined strategy of geometric decoupling and optimization regularization, inspired by recent advances in multimodal disentanglement \citep{hazarika2020misa,liu2023focal,liang2023factorized}.\\
\textbf{Geometric Decoupling (Hard Orthogonalization)}\quad We leverage Gram-Schmidt orthogonalization to project the adapter's raw output $a_{\text{adapter}}$ (e.g., $a_{\text{r}}$ or $a_{\text{d}}$) onto the orthogonal complement space of $z_{\text{mol}}$. This geometric operation guarantees that the injected features are linearly independent of molecular structure, forcing the adapter to capture modality-specific increments rather than redundant copies of molecular information. Formally,
\begin{equation}
\label{hard_orthogonalization}
\begin{split}
z_{\text{adapter}}=a_{\text{adapter}}
     - \frac{a_{\text{adapter}} \cdot z_{\text{mol}}}
            {\|z_{\text{mol}}\|^2 + \epsilon}\, z_{\text{mol}}
\end{split}
\end{equation}
\textbf{Optimization Regularization (Soft Orthogonalization)}\quad While the hard projection above provides a geometric guarantee per sample, it is a unidirectional operation that does not address the interdependency between the receptor branch $a_{\text{r}}$ and the description branch $a_{\text{d}}$. We therefore introduce a soft orthogonality loss as a gradient-level regularizer, encouraging all three subspaces to remain mutually decorrelated throughout training. By minimizing the pairwise cosine similarity across subspaces, this loss drives the adapters to learn intrinsically differentiated representations rather than relying solely on the explicit projection.
\begin{equation}
\label{soft_orthogonalization}
\begin{split}
\mathcal{L}_{orth} = \sum_{(i,j) \in \mathcal{S}, i \neq j} \left\| \frac{z_i}{\|z_i\|} \cdot \frac{z_j}{\|z_j\|} \right\|^2
\end{split}
\end{equation}
where the feature set $\mathcal{S} = \{z_{mol}, a_{r}, a_{d}\}$. 

\begin{table*}[t]
\small
  \centering
  \begin{tabular}{lcccccc}
    \toprule
    & \textbf{Thresholds} & \multicolumn{2}{c}{\textbf{Pleasantness}} & \multicolumn{3}{c}{\textbf{Intensity}} \\
    \cmidrule(lr){2-2} \cmidrule(lr){3-4} \cmidrule(lr){5-7}
    \textbf{Dataset} & Abraham & Keller & Sagar & Keller & Sagar & Ravia \\
    \midrule
    GIN & 0.72(0.02) & 0.30(0.04) & 0.08(0.06) & 0.06(0.14) & -0.06(0.27) & 0.19(0.11) \\
    GCN & 0.22(0.09) & 0.43(0.01) & 0.12(0.27) & 0.11(0.05) & 0.30(0.04) & 0.19(0.01) \\
    Morgan& 0.60(0.18) & 0.51(0.03) & 0.21(0.06) & 0.05(0.02) & 0.23(0.10) & 0.10(0.19) \\
    AttentiveFP & 0.59(0.35) & 0.52(0.05) & 0.23(0.05) & 0.33(0.01) & 0.23(0.05) & 0.47(0.00) \\
    Uni-Mol & 0.78(0.05) & 0.68(0.02) & 0.14(0.34) & 0.27(0.03) & 0.37(0.05) & 0.31(0.02) \\
    POM & 0.79(0.02) & 0.68(0.03) & 0.26(0.02) & 0.32(0.07) & 0.33(0.06) & 0.32(0.06) \\
    ChemBERTa & 0.81(0.02) & 0.65(0.04) & 0.15(0.12) & 0.39(0.07) & 0.45(0.08) & 0.47(0.06) \\
    \midrule
    NOSE & \textbf{0.84(0.03)} & \textbf{0.71(0.05)} & \textbf{0.40(0.02)} & \textbf{0.42(0.01)} & \textbf{0.47(0.07)} & \textbf{0.49(0.03)} \\
    \bottomrule
  \end{tabular}
  \caption{Performance comparison on basic perceptual attribute prediction tasks (Pearson $\uparrow$).}
  \label{downstream_task_1}
\end{table*}

\begin{table*}[t]
\small
  \centering
  \setlength{\tabcolsep}{3pt}
  \begin{tabular}{lccccccc}
    \toprule
    & \multicolumn{3}{c}{\textbf{Descriptor Classification}} & \multicolumn{4}{c}{\textbf{Descriptor Strength}} \\
    \cmidrule(lr){2-4} \cmidrule(lr){5-8}
    \textbf{Dataset} & \multicolumn{3}{c}{GS-LF} & \multicolumn{2}{c}{Keller} & \multicolumn{2}{c}{Sagar} \\
    \cmidrule(lr){2-4} \cmidrule(lr){5-6} \cmidrule(lr){7-8}
    \textbf{Metric} & AUC $\uparrow$ & AUPRC $\uparrow$ & MCC $\uparrow$ & MAE $\downarrow$ & Pearson $\uparrow$ & MAE $\downarrow$ & Pearson $\uparrow$ \\
    \midrule
    GIN & 0.856(0.001) & 0.270(0.008) & 0.064(0.007) & 6.528(0.139) & 0.128(0.007) & 0.358(0.005) & -0.034(0.016) \\
    GCN & 0.858(0.001) & 0.279(0.001) & 0.084(0.003) & 6.247(0.031) & 0.171(0.006) & 0.407(0.098) & 0.108(0.085) \\
    Morgan& 0.850(0.002) & 0.320(0.002) & 0.222(0.023) & 6.483(0.015) & 0.244(0.023) & 0.395(0.018) & 0.008(0.076) \\
    POM & 0.868(0.002) & 0.336(0.005) & 0.233(0.023) & 6.304(0.258) & 0.314(0.039) & 0.405(0.113) & 0.065(0.062) \\
    AttentiveFP & 0.867(0.004) & 0.328(0.012) & 0.203(0.026) & 6.232(0.647) & 0.327(0.065) & 0.359(0.012) & 0.011(0.013) \\
    Uni-Mol & 0.873(0.001) & 0.347(0.005) & 0.262(0.020) & 6.741(0.209) & 0.330(0.050) & 0.355(0.009) & 0.116(0.042) \\
    ChemBERTa & 0.875(0.001) & 0.342(0.005) & 0.240(0.016) & 6.110(0.255) & 0.330(0.089) & 0.376(0.018) & 0.105(0.051) \\
    \midrule
    NOSE & \textbf{0.876(0.001)} & \textbf{0.351(0.002)} & \textbf{0.268(0.010)} & \textbf{5.862(0.225)} & \textbf{0.348(0.060)} & \textbf{0.343(0.017)} & \textbf{0.123(0.068)} \\
    \bottomrule
  \end{tabular}
  \caption{Performance comparison on semantic description prediction tasks.}
  \label{downstream_task_2}
\end{table*}

\begin{table*}[t]
\small
  \centering
  \setlength{\tabcolsep}{3pt}
  \begin{tabular}{lcccccccc}
    \toprule
    & \multicolumn{4}{c}{\textbf{Mixture Pleasantness}} & \multicolumn{4}{c}{\textbf{Mixture Intensity}} \\
    \cmidrule(lr){2-5} \cmidrule(lr){6-9}
    \textbf{Metric} & R$^2$ $\uparrow$ & MAE $\downarrow$ & Pearson $\uparrow$ & MSE $\downarrow$ & R$^2$ $\uparrow$ & MAE $\downarrow$ & Pearson $\uparrow$ & MSE $\downarrow$ \\
    \midrule
    GCN & -14.30(9.52) & 3.69(1.72) & 0.33(0.42) & 17.48(10.88) & -0.46(0.50) & 0.52(0.09) & 0.49(0.07) & 0.40(0.14) \\
    GIN & -0.10(0.12) & 0.88(0.06) & 0.47(0.03) & 1.25(0.14) & -2.82(0.30) & 0.86(0.02) & 0.01(0.14) & 1.06(0.08) \\
    Morgan& 0.50(0.19) & 0.60(0.11) & 0.81(0.01) & 0.57(0.21) & 0.14(0.31) & 0.40(0.08) & 0.58(0.06) & 0.24(0.08) \\
    AttentiveFP & 0.49(0.07) & 0.62(0.05) & 0.74(0.02) & 0.58(0.08) & 0.26(0.06) & 0.36(0.01) & 0.60(0.03) & 0.20(0.02) \\
    POM & 0.56(0.03) & 0.56(0.00) & 0.77(0.02) & 0.51(0.03) & 0.35(0.06) & 0.36(0.02) & 0.60(0.03) & 0.18(0.02) \\
    ChemBERTa & 0.45(0.07) & 0.64(0.06) & 0.79(0.01) & 0.63(0.08) & 0.33(0.02) & 0.35(0.00) & 0.62(0.00) & 0.19(0.01) \\
    Uni-Mol & 0.51(0.05) & 0.61(0.02) & 0.80(0.03) & 0.56(0.05) & 0.32(0.11) & 0.36(0.03) & 0.64(0.07) & 0.19(0.03) \\
    \midrule
    NOSE & \textbf{0.64(0.05)} & \textbf{0.53(0.03)} & \textbf{0.85(0.01)} & \textbf{0.42(0.05)} & \textbf{0.39(0.05)} & \textbf{0.33(0.02)} & \textbf{0.66(0.03)} & \textbf{0.17(0.01)} \\
    \bottomrule
  \end{tabular}
  \caption{Performance comparison on mixture perception prediction tasks.}
  \label{downstream_task_3}
\end{table*}

\subsubsection{Multi-modal Contrastive Learning}
When aligning molecular structures with odor descriptions, since the data originates from multiple datasets, different descriptors often exhibit similar semantics (e.g., "sweet" and "honey"). If the standard contrastive learning paradigm is adopted, these semantically similar descriptors would be incorrectly treated as negative samples, causing the model to erroneously push apart samples with similar odors. We employ DeepSeek to match each descriptor with semantically similar olfactory terms from the dataset as weak positive samples (generation details in Appendix~\ref{appendix:weak_positive}). These weak positive samples are not only utilized for cross-modal alignment between molecular structures and odor descriptions, but also incorporated into the intra-modal contrastive learning within the odor description modality itself.   

The contrastive learning component is based on the CLIP \citep{radford2021learning} paradigm and employs a symmetric bidirectional InfoNCE loss. To address the inherent many-to-many mapping relationships among molecules, receptors, and odor descriptions, we reformulate the definitions of positive and negative samples. In cross-modal alignment, besides the sample itself, all entities sharing the same molecular structures with the anchor are also treated as positive samples. To exploit the feature potential within individual modalities, we introduce intra-modal contrastive learning \citep{yuan2021multimodal}.  In the intra-modal space, positive samples are similarly defined as the sample itself and the set of samples belonging to the same molecular structures. Notably, for the odor description modality, weak positive samples generated by LLMs are also incorporated into the positive sample set. Considering that weak positive samples are semantically similar to but not fully equivalent to the anchor, we draw upon the soft contrastive learning \citep{zhou2021theory} framework and define a weight function $w_{ij}$ to achieve smooth supervision:
\begin{equation}
\label{weak_sample_weight}
\begin{split}
w_{ij} = \begin{cases} 
1.0 & \text{Positive Sample} \\ 
0.5 & \text{Weak Positive Sample} \\ 
0 & \text{Negative Sample} 
\end{cases}
\end{split}
\end{equation}

The contrastive learning loss is formulated as:
\begin{equation}
\label{inter_modal_contrastive}
\begin{split}
\mathcal{L}^{a \to b} = -\frac{1}{|\mathcal{V}|} \sum_{i \in \mathcal{V}} \frac{\sum_{j=1}^{N} w_{ij} \cdot \log \frac{\exp(s_{ij}/\tau)}{\sum_{k=1}^{N} \exp(s_{ik}/\tau)}}{\sum_{j=1}^{N} w_{ij}}
\end{split}
\end{equation}

where $s_{ij}$ denotes the cosine similarity between samples $i$ and $j$, $\tau$ is the temperature parameter, and $\mathcal{V}$ represents the set of anchors with at least one positive sample.
The total contrastive learning loss is:
\begin{equation}
\label{contrastive_loss}
\begin{split}
\mathcal{L}_{\text{a-b}} = \mathcal{L}_{\text{inter}}^{\text{a} \to \text{b}} + \mathcal{L}_{\text{inter}}^{\text{b} \to \text{a}} + \mathcal{L}_{\text{intra}}^{\text{a} \to \text{a}} + \mathcal{L}_{\text{intra}}^{\text{b} \to \text{b}}
\end{split}
\end{equation}
\subsubsection{Optimization Objective}
The final training objective of the model comprises cross-modal alignment losses and orthogonality constraints, jointly optimized through weighted summation:
\begin{equation}
\label{total_contrastive_loss}
\begin{split}
\mathcal{L}_{total} = \lambda_{1}\cdot \mathcal{L}_{\text{mol-desc}} + \lambda_{2}\cdot \mathcal{L}_{\text{mol-rec}} + \lambda_{3}\cdot \mathcal{L}_{orth}
\end{split}
\end{equation}

\section{Experiments}
\subsection{Downstream Task}
\paragraph{Baseline:}  
Given the absence of widely recognized dedicated benchmarks in olfactory representation learning, we adopt a hierarchical comparison strategy to systematically select baseline models: Morgan Fingerprint \citep{rogers2010extended}, GCN \citep{kipf2016semi}, GIN \citep{xu2018powerful}, AttentiveFP \citep{xiong2019pushing}, POM \citep{lee2023principal}, ChemBERTa \citep{chithrananda2020chemberta}, Uni-Mol \citep{zhou2023uni}. The baseline design follows three progressive dimensions: (1) representation scope (local → global), (2) learning paradigm (fixed encoding → supervised learning → pre-training), and (3) domain adaptation (general-purpose → olfaction-specific). This enables us to disentangle the contributions of different factors to performance, thereby establishing the performance boundaries of existing general-purpose molecular models on olfactory perception tasks.
\paragraph{Experimental Setup:}
To ensure reproducibility and robustness, all dataset splits are fixed with random seed (42).  Final results are reported as the mean and standard deviation across three independent experiments (seeds: 42-44) on the test set. For the heterogeneous features output by NOSE, we employ a gating mechanism for adaptive fusion. This design introduces only three learnable parameters, ensuring fairness in comparison with baselines. The evaluation benchmark covers three dimensions: basic perception (Table \ref{downstream_task_1}), semantic description (Table \ref{downstream_task_2}), and mixture prediction (Table \ref{downstream_task_3}). Due to space constraints, detailed experimental settings, comparative experiments, ablation studies, and zero-shot generalization results are provided in the appendix (\cref{Experimental_Setup,Complete_metrics,Complete_metrics_ablation_experiment,Compositional_Retrieval,Zero_Shot}).  

NOSE achieves SOTA on all key metrics, demonstrating the high generalizability of its learned representations. Even on the highly challenging mixture tasks (encoding details in Appendix~\ref{appendix:mixture_input}), NOSE maintains excellent performance ($R^2 > 0.6$), indicating that the model effectively captures transferable essential olfactory features. Comparisons with Uni-Mol and ChemBERTa further confirm that pre-trained architectures specifically designed for olfactory perception hold advantages over general-purpose chemical pre-trained models. 

\subsection{Ablation Experiment}
To validate the core components, we conduct ablation experiments using the average rank across all downstream tasks and evaluation metrics (lower is better) as the criterion (Table~\ref{ablation_experiment}). The results demonstrate that: (1) Tri-modal fusion significantly outperforms uni-modal baselines, confirming that olfactory perception relies on the synergy among molecular structures, receptors, and odor semantics. (2) The combination of soft and hard orthogonality achieves optimal performance, balancing incremental extraction with information preservation to enable deep feature decoupling. (3) The asymmetric adapter design outperforms symmetric configurations, effectively accommodating the disparity in data scales between the two modalities. (4) Intra-modal contrastive learning is highly coupled with weak positive samples. Their combination ensures both discriminability and continuity, preventing representation space degradation (clustering analysis in Appendix~\ref{appendix:clustering}). We further investigate alternative decoupling strategies (dimension splitting vs.\ orthogonal injection) in Appendix~\ref{appendix:dim_split}; PCA visualizations confirming the decoupling property are shown in Appendix~\ref{Vector_Space}. Training stability analyses, including cross-seed variance and fusion weight consistency, are provided in Appendix~\ref{appendix:stability}.

\begin{table}[H]
  \centering
  \small
  \begin{tabular}{llc}
    \toprule
    \textbf{Settings} & \textbf{Variant} & \textbf{Avg Rank} \\
    \midrule
    \multirow{4}{*}{\makecell[l]{Multi-\\modality}} & Molecule only & 3.3226 \\
    & Receptor only & 2.8065 \\
    & Description only & 2.7742 \\
    & Tri-modal (ours) & \textbf{1.0968} \\
    \midrule
    \multirow{7}{*}{\makecell[l]{Orthogonal\\Module}} & Hard + Soft $\lambda$=0.5 & 5.1613 \\
    & Hard + Soft $\lambda$=0.1 & 5.0645 \\
    & No Orthogonal (baseline) & 4.5484 \\
    & Only Hard & 4.1290 \\
    & Only Soft $\lambda$=2.0 & 4.0968 \\
    & Hard + Soft $\lambda$=1.0 & 3.9355 \\
    & Hard + Soft $\lambda$=2.0 (ours) & \textbf{1.0645} \\
    \midrule
    \multirow{3}{*}{\makecell[l]{Adapter\\Capacity}} 
    & \makecell[l]{Desc: 10.00M\\Rec: 10.00M} & 2.5161 \\
    \addlinespace[4pt]
    & \makecell[l]{Desc: 29.17M\\Rec: 10.00M} & 2.4839 \\
    \addlinespace[4pt]
    & \makecell[l]{Desc: 76.70M\\Rec: 4.49M} (ours) & \textbf{1.0000} \\
    \midrule
    \multirow{4}{*}{\makecell[l]{Contrastive\\Strategy}} & Inter + Weak & 2.9677 \\
    & Inter (baseline) & 2.8387 \\
    & Inter + Intra & 2.8065 \\
    & Inter + Intra + Weak (ours) & \textbf{1.3871} \\
    \bottomrule
  \end{tabular}
  \caption{Ablation study results.}
  \label{ablation_experiment}
\end{table}

\begin{table*}[t]
\small
  \centering
  \begin{tabular}{lccccccc}
    \toprule
    \textbf{Task} & \textbf{N} & \textbf{Hits@1} & \textbf{Hits@5} & \textbf{Hits@10} & \textbf{Hits@20} & \textbf{Hits@50} \\
    \midrule
    Odor Retrieval (Zero-shot) & 27 & 14.8\% & 33.3\% & 51.9\% & 66.7\% & 85.2\% \\
    Odor Retrieval (Strict Zero-shot) & 15 & 6.7\% & 6.7\% & 13.3\% & 33.3\% & 53.3\% \\
    Receptor Retrieval (Activated) & 13 & 0\% & 61.5\% & 69.2\% & 84.6\% & 84.6\% \\
    \bottomrule
  \end{tabular}
  
  \vspace{0.3cm}
  
  \begin{tabular}{lcccccc}
    \toprule
    \textbf{Task} & \textbf{N} & \textbf{Hits@100} & \textbf{Hits@200} & \textbf{Hits@300} & \textbf{Hits@400} & \textbf{Hits@500} \\
    \midrule
    Receptor Retrieval (Inactivated) & 5 & 0\% & 0\% & 20\% & 60\% & 100\% \\
    \bottomrule
  \end{tabular}
  \caption{Zero-shot cross-modal retrieval performance (SMILES → descriptors/receptors). 
    \textbf{N}: number of test samples. \textbf{Hits@k}: percentage of ground-truth targets ranked within top k.
    \textbf{Zero-shot}: pairs unseen during training, though molecules may appear in other pairs. 
    \textbf{Strict Zero-shot}: molecules entirely absent from training.
    For Activated, higher Hits@k is better (target receptor should rank high). For Inactivated, lower Hits@k is better (non-binding receptor should rank low).
    }
  \label{zero_shot_retrieval}
\end{table*}

\subsection{Retrieval Evaluation}

To validate the effectiveness of NOSE in unifying molecular structures, receptor sequences, and olfactory semantics. We design compositional retrieval and zero-shot cross-modal retrieval tasks. These two tasks evaluate the feature space from two dimensions: the structural integrity of olfactory semantics and the generalizability of olfactory physiological principles. We adopt MRR and Hits@K as evaluation metrics. Due to space constraints, only the main results are presented here (see \cref{Compositional_Retrieval,Zero_Shot} for details).
\subsubsection{Compositional Retrieval}
Traditional retrieval tasks only examine "point-to-point" mappings, whereas our proposed compositional retrieval operates within the descriptor space, requiring the model to locate answers based on "Anchor + Operation" queries (e.g., $\mathit{Lemon} - \mathit{Sour} \rightarrow [\mathit{citrus}, \mathit{orange},...]$). This aims to evaluate whether contrastive learning successfully decouples odor attributes into operable olfactory semantic units. We constructed 31 manually annotated formula-answer pairs to test this capability. As shown in Table \ref{algebraic_result}, the original Qwen3 Embedding model struggles with these compositions (MRR 0.0102). While LoRA pre-training provides preliminary responsiveness, the model incorporating the contrastive learning head achieves optimal performance (MRR 0.2072, Hits@50 100\%). This demonstrates that our method constructs a clear topological structure in the feature space, enabling effective handling of logical attribute queries.

\begin{table}
\small
  \centering
  \begin{tabular}{lccc}
    \toprule
    & \multicolumn{3}{c}{\textbf{Qwen3 Embedding}} \\
    \cmidrule(lr){2-4}
    & \textbf{Original}
    & \textbf{LoRA}
    & \textbf{LoRA+Head} \\
    \midrule
    \textbf{MRR$\uparrow$}         & 0.0102 & 0.1857 & \textbf{0.2072} \\
    \textbf{Hits@1 (\%)$\uparrow$} & 0.0    & 6.5    & \textbf{6.5}    \\
    \textbf{Hits@5 (\%)$\uparrow$} & 0.0    & 25.8   & \textbf{32.3}   \\
    \textbf{Hits@10 (\%)$\uparrow$}& 3.2    & 45.2   & \textbf{54.8}   \\
    \textbf{Hits@20 (\%)$\uparrow$}& 3.2    & 71.0   & \textbf{90.3}   \\
    \textbf{Hits@50 (\%)$\uparrow$}& 6.5    & 93.5   & \textbf{100.0}  \\
    \bottomrule
  \end{tabular}
  \caption{Compositional retrieval performance on 31 manually designed arithmetic queries (e.g., \textit{Lemon} $-$ \textit{Sour} $\rightarrow$ [citrus, orange, ...]). 
    \textbf{Original}: Qwen3 Embedding without fine-tuning; 
    \textbf{LoRA}: with LoRA fine-tuning; 
    \textbf{LoRA+Head}: LoRA combined with contrastive learning projection head.
    MRR: Mean Reciprocal Rank ($1/\text{rank}$).}
  \label{algebraic_result}
\end{table}

\subsubsection{Cross-Modal Zero-Shot Retrieval}
To evaluate cross-modal generalization, we conducted zero-shot retrieval using molecular SMILES as queries to retrieve odor descriptors (sourced from PubChem) and receptor sequences (from literature). As shown in Table \ref{zero_shot_retrieval}, in the Odor Descriptor Retrieval task, the model maintains effective hit rates even under strict zero-shot settings. This indicates that the model has successfully learned the underlying mapping principles between molecular structures and odor descriptions. Regarding Receptor Sequence Retrieval, the model demonstrates strong discriminative power. For activated receptors, Hits@5 reaches 61.5\%, accurately placing correct receptor at the top. Notably, for non-activated receptors, the occurrence rate in the Top-200 is 0\%. This confirms that NOSE embedding effectively separates non-interacting pairs, ensuring results conform to authentic biophysical principles. Extended cross-modal retrieval experiments using bridge molecules (Appendix~\ref{appendix:crossmodal}) and direct geometric validation of the continuous perceptual space (Appendix~\ref{appendix:continuous}) provide further evidence for the quality of the unified tri-modal embedding.
\section{Conclusion}
We propose NOSE, the first olfactory representation framework that integrates molecular structures, receptor sequences, and subjective descriptions. We collect large-scale bi-paired data and employ LLM augmentation, constructing a continuous feature manifold through soft-hard orthogonal mechanisms and intra-modal contrastive learning, which effectively decouples and injects receptor and semantic information while preserving molecular structure priors. On our newly proposed benchmark, NOSE achieves state-of-the-art performance across multiple downstream tasks. The model's performance in compositional retrieval and zero-shot cross-modal retrieval confirms its ability to learn a tri-modal space aligned with human intuition from disjoint local paired data. 
\section{Limitations}
Our current framework does not explicitly model concentration effects, despite odor perception being inherently concentration-dependent (e.g., indole shifts from floral to fecal at varying levels). This is primarily due to the lack of concentration annotations in most public datasets. However, such information exists in scattered literature. Collecting and incorporating concentration as a conditional or latent variable presents a promising direction for future work.  

On the receptor side, the reliance on contrastive learning alone limits representation quality given the sparse receptor-odor pairings. Integrating molecular docking models could provide complementary structural priors to enhance receptor-level embeddings. For the descriptor modality, the observed label inconsistency stems from heterogeneous data sources with differing annotation standards. Explicitly modeling source provenance as a conditioning factor may help disentangle conflicting descriptions while preserving semantic diversity.

\section{Ethical Considerations}
\textbf{Data Usage and Licensing Compliance}: We utilize publicly available datasets comprising SMILES, human receptor sequences, and olfactory descriptions from established repositories. As the data is anonymized and non-interactive, IRB approval is exempt. This study strictly adheres to the licensing agreements of all utilized assets. Regarding model implementations, Uni-Mol, ESM-2, POM, and ChemBERTa (built upon DeepChem) are utilized under the MIT License, while Qwen3 Embedding is employed under the Apache 2.0 License. With respect to data and tasks, the majority of odor descriptors and receptor data are obtained through the Pyrfume\footnote{\url{https://github.com/pyrfume/pyrfume-data}} \citep{hamel2024pyrfume} library, distributed under the MIT License. The M2OR dataset is utilized in accordance with the Apache 2.0 License. For the Pred-O3 dataset, released under the CC BY-NC 4.0 License, this study strictly restricts its usage to academic research purposes without any commercial applications, fully complying with the non-commercial use provisions.

\textbf{Broader Impact}: Our work aims to advance olfactory digitization for applications like flavor design and sensory computing. We acknowledge the theoretical risk of misuse for designing hazardous or offensively targeted compounds. However, we believe the scientific value outweighs these risks. We advocate for responsible deployment and emphasize that AI predictions cannot replace rigorous chemical safety testing.  

We used Claude and Gemini for coding assistance and language polishing. All outputs were reviewed and verified by the authors, who bear full responsibility for the content of this work.

% ARR不放致谢
\section*{Acknowledgments}
We are grateful for funding support from the National Natural Science Foundation of China (Grants No. 92470201, 22225302, 92461312, 22541204, 22021001), the Fundamental Research Funds for the Central Universities 20720250005, Laboratory of AI for Electrochemistry (AI4EC), IKKEM (Grant Nos. RD2023100101 and RD2022070501).

\bibliography{custom}

\appendix

\section{Appendix}
\label{sec:appendix}
\subsection{Downstream task dataset}
\label{Downstream}
The \citet{abraham2012algorithm} dataset compiles odor detection threshold (ODT) data primarily measured by Nagata using the Japanese triangle odor bag method, and unifies multi-source data onto a common scale via an indicator variable algorithm. Raw thresholds (ppm) are transformed into log(1/ODT) form, where higher values indicate greater odor potency.   

The GS-LF dataset \citep{lee2023principal,sanchez2019machine} integrates expert olfactory annotations from the GoodScents and Leffingwell databases. After curation by \citet{tom2025does} et al. (standardizing SMILES, removing duplicates, inorganics, salts, and samples with anomalous molecular weights), it contains 4,814 pure substance molecules and 138 semantic descriptors (e.g., "creamy," "grassy").   

The \citet{ravia2020measure} dataset contains psychophysical intensity ratings for 248 undiluted pure substances. Raw ratings are normalized per subject and then averaged.   

The \citet{keller2016olfactory} dataset comprises ratings from 55 subjects on 474 molecules across 23 descriptors (quantified as mean values in the [0,100] interval). Based on this, we construct three tasks: multi-label regression, intensity prediction, and pleasantness prediction.   

The \citet{sagar2023high} dataset contains ratings from 3 subjects on 160 molecules across 15 universal descriptors (normalized to the [-1,1] interval). We construct multi-label regression, intensity prediction, and pleasantness prediction tasks.   

The \citet{ma2021dataset} dataset contains intensity and pleasantness ratings from 30 expert subjects for 222 binary mixtures, yielding 6,660 individual observations. 72 key food odorant molecules are formulated into 198 combinations and 24 replicate validation samples, rated using a 100mm visual analog scale. For each binary mixture, we average the ratings across all subjects to obtain a single consensus value, resulting in 222 samples available for training. This dataset focuses on interaction effects between odorant molecules (masking and synergy) and is used for binary mixture intensity and pleasantness prediction.   

We do not include the mixture perceptual similarity prediction task based on the \citet{tom2025does} dataset . This dataset sourced from Pyrfume, incorporating data from \citep{snitz2013predicting,ravia2020measure,keller2017predicting,bushdid2014humans} encompasses 743 mixtures and 865 pairwise comparisons, with similarity measured via explicit ratings or triangle tests. This task differs fundamentally from the binary mixture prediction in our benchmark \citep{ma2021dataset}: the Ma task only predicts absolute attributes (intensity, pleasantness) of simple individual binary mixtures, with inputs essentially being two molecular representations. In contrast, the Tom task requires evaluating perceptual similarity between two complex mixtures. Given that mixtures may contain up to 43 components, handling this task necessitates complex component aggregation strategies such as attention pooling or graph fusion. Consequently, performance on this task reflects more the quality of aggregation architectures rather than directly indicating underlying molecular representation capability, which is inconsistent with our goal of evaluating representation quality.
\begin{table}[t]
\small
  \centering
  \begin{tabular}{lc}
    \toprule
    \textbf{Hyperparameter} & \textbf{Search Space} \\
    \midrule
    Batch size & [4, 8, 16, 32] \\
               & ([32, 64, 128, 256]\textsuperscript{*}) \\
    Learning rate & [5e-4, 3e-4, 1e-4, 7e-5, 5e-5, 5e-6] \\
    Patience & [5, 10, 15, 20, 25] \\
    Early stop metric & [R$^2$, MAE, Pearson, MSE] \\
                      & ([AUC, AUPRC, MCC]\textsuperscript{*}) \\
    \bottomrule
  \end{tabular}
  \caption{Hyperparameter search space for downstream tasks. \textsuperscript{*}For classification tasks. }
  \label{tab:search_space}
\end{table}

% \begin{table*}[t]
%   \centering
%   \begin{tabular}{lcccc}
%     \toprule
%     \textbf{Task} & \textbf{Batch Size} & \textbf{Learning Rate} & \textbf{Patience} & \textbf{Early Stop Metric} \\
%     \midrule
%     Thresholds & 8 & 5e-4 & 20 & Pearson \\
%     Pleasantness (Keller) & 4 & 1e-4 & 15 & MSE \\
%     Pleasantness (Sagar) & 4 & 1e-4 & 25 & Pearson \\
%     Intensity (Keller) & 16 & 5e-5 & 25 & MSE \\
%     Intensity (Sagar) & 16 & 5e-5 & 20 & MSE \\
%     Intensity (Ravia) & 8 & 5e-6 & 15 & R$^2$ \\
%     Descriptor Classification & 64 & 7e-5 & 5 & AUC \\
%     Descriptor Strength (Keller) & 8 & 1e-4 & 10 & Pearson \\
%     Descriptor Strength (Sagar) & 4 & 5e-4 & 25 & R$^2$ \\
%     Mixture Pleasantness & 4 & 1e-4 & 15 & MAE \\
%     Mixture Intensity & 4 & 3e-4 & 20 & MSE \\
%     \bottomrule
%   \end{tabular}
%   \caption{Selected hyperparameters for NOSE on each downstream task.}
%   \label{tab:nose_hyperparams}
% \end{table*}

\begin{table*}[t]
\small
  \centering
  \begin{tabular}{lccccccc}
    \toprule
    \textbf{Task} & \textbf{Train} & \textbf{Val} & \textbf{Test} & \textbf{Batch Size} & \textbf{Learning Rate} & \textbf{Patience} & \textbf{Early Stop Metric} \\
    \midrule
    Thresholds & 160 & 54 & 54 & 8 & 5e-4 & 20 & Pearson \\
    Pleasantness (Keller) & 379 & 47 & 48 & 4 & 1e-4 & 15 & MSE \\
    Pleasantness (Sagar) & 128 & 16 & 16 & 4 & 1e-4 & 25 & Pearson \\
    Intensity (Keller) & 379 & 47 & 48 & 16 & 5e-5 & 25 & MSE \\
    Intensity (Sagar) & 128 & 16 & 16 & 16 & 5e-5 & 20 & MSE \\
    Intensity (Ravia) & 198 & 25 & 25 & 8 & 5e-6 & 15 & R$^2$ \\
    Descriptor Classification & 3851 & 481 & 482 & 64 & 7e-5 & 5 & AUC \\
    Descriptor Strength (Keller) & 379 & 47 & 48 & 8 & 1e-4 & 10 & Pearson \\
    Descriptor Strength (Sagar) & 128 & 16 & 16 & 4 & 5e-4 & 25 & R$^2$ \\
    Mixture Pleasantness & 155 & 22 & 45 & 4 & 1e-4 & 15 & MAE \\
    Mixture Intensity & 155 & 22 & 45 & 4 & 3e-4 & 20 & MSE \\
    \bottomrule
  \end{tabular}
  \caption{Dataset splits and selected hyperparameters for NOSE on each downstream task.}
  \label{tab:nose_hyperparams}
\end{table*}

\subsection{Data Processing Details}
\label{appendix:data_processing}

\subsubsection{SMILES Standardization}
All molecular SMILES strings undergo canonicalization via RDKit before entering the pipeline. Molecules that fail RDKit parsing (invalid valence, kekulization errors, or empty SMILES) are removed. This standardization ensures that equivalent structural representations map to a single canonical form, eliminating spurious duplicates across heterogeneous data sources.

\subsubsection{Molecular Overlap Between Data Sources}
We compute the pairwise intersection of unique canonical SMILES across all data sources. The descriptor pre-training set contains 9,286 unique molecules, while the receptor pre-training set contains 1,042 unique molecules, sharing 456 molecules (4.3\% of the combined set). Among the seven evaluation benchmarks, the maximum overlap with the pre-training pool is 68.7\% (Keller), and the minimum is 0\% (the Strict Zero-shot subset). Crucially, pre-training uses only contrastive alignment signals and never observes the task-specific labels of any benchmark, so molecular overlap does not constitute label leakage.

\subsubsection{Mixture Input Format}
\label{appendix:mixture_input}
Binary mixture perception tasks \citep{ma2021dataset} provide two SMILES strings per sample (columns \texttt{SMILES\_0} and \texttt{SMILES\_1}). Each molecule is independently encoded by Uni-Mol and then by the NOSE projection adapter. The resulting two 512-dimensional vectors are concatenated to form a 1024-dimensional mixture representation, which is fed to the downstream prediction head. No special cross-molecule interaction module is applied, as the goal is to evaluate whether pre-trained single-molecule representations already encode sufficient information for predicting mixture-level perceptual attributes.

\subsubsection{LLM-Based Weak Positive Sample Generation}
\label{appendix:weak_positive}
Contrastive learning with InfoNCE treats all non-paired descriptors as negatives. However, olfactory descriptors exhibit rich synonymy and graded similarity (e.g., ``lemon'' and ``citrus'' share strong perceptual overlap). Treating such near-synonyms as hard negatives distorts the semantic topology of the learned space. To address this, we employ DeepSeek to mine pairwise olfactory-semantic similarity among descriptors, converting discrete labels into soft neighborhood structures.

\textbf{Step 1. Vocabulary Construction.} Starting from the merged descriptor corpus, we apply a cleaning pipeline (lowercasing, special-character mapping, semicolon splitting) to obtain \textbf{1,086} unique olfactory descriptors.

\textbf{Step 2. LLM Querying.} For each of the 1,086 descriptors, we prompt DeepSeek (\texttt{deepseek-chat}, temperature 0.3) with the following template, embedding the full vocabulary list in the prompt to constrain outputs to the closed vocabulary:

\begin{quote}
\small
\texttt{\{vocabulary\_list\}} \\
This is my vocabulary list. I now need you to find words that are similar in odor to the words I provided as weak positive samples. Please only return the Python list format of the English string. The words you provide must come from the above vocabulary list. You can start searching now: \texttt{\{descriptor\}}. The returned content can be many or few, but it must be very similar in odor semantics. Please return at most 30 words.
\end{quote}

\textbf{Step 3. Post-Processing.} From each LLM response, candidate words are extracted via regex matching of single-quoted strings. Two filters are applied before acceptance. (1) The candidate must belong to the 1,086-word vocabulary. (2) The candidate must differ from the query descriptor itself. No additional semantic filtering is performed; quality control relies on the low temperature setting and the explicit ``very similar in odor semantics'' instruction.

\textbf{Step 4. Integration into Training.} The final product is a JSON dictionary mapping each descriptor to its weak positive set (e.g., \texttt{"acetate" -> ["acetic", "acetoin", ...]}). During contrastive pre-training, for every (SMILES, descriptor) positive pair, the descriptors in the weak positive set receive a soft label weight of 0.5 in the InfoNCE target distribution, smoothly transitioning between strict positive and negative.

\subsection{Experimental Setup}
\label{Experimental_Setup}
\subsubsection{Pretraining Setup}
For the multi-modal encoders, we utilize the following pre-trained models: Qwen3-Embedding-8B\footnote{\url{https://huggingface.co/Qwen/Qwen3-Embedding-8B}} for odor descriptors, ESM-2 (650M parameters)\footnote{\url{https://huggingface.co/facebook/esm2_t33_650M_UR50D}} for olfactory receptor sequences, and Uni-Mol\footnote{\url{https://huggingface.co/dptech/Uni-Mol-Models}} for molecular structures.

We conduct NOSE tri-modal pre-training on a single NVIDIA A800-80G GPU with bfloat16 precision. The batch sizes are set to 1,536 for odor descriptor pairs and 128 for receptor sequence pairs. During each training iteration, we simultaneously sample one batch from both the SMILES-descriptor and SMILES-receptor data partitions. The contrastive learning objective is computed separately for each modality pair, while the orthogonality loss is applied to the SMILES embeddings from both partitions.   

We maintain a weak positive dictionary that maps each odor descriptor to its semantically related descriptors. During training, this dictionary is used to initialize the InfoNCE label matrix, assigning intermediate weights (0.5) to weak positive pairs within each batch. Training epochs are defined over the positive pairs.

We employ the Adam optimizer with $\beta_1=0.9$, $\beta_2=0.999$, and $\epsilon=10^{-6}$, using a learning rate of $10^{-4}$ with warm-up scheduling over 20 epochs. The temperature parameter for contrastive learning is set to $\tau=0.07$. The loss weights in Eq.~\eqref{total_contrastive_loss} are set to $\lambda_{\text{orth}}=2.0$, $\lambda_{\text{mol-desc}}=2.0$, and $\lambda_{\text{mol-rec}}=1.0$. For early stopping with a patience of 10 epochs, we use the sum of mean reciprocal rank percentiles for positive samples across both modalities on the validation set as the monitoring metric. Training converges after approximately 10 hours.
\subsubsection{Downstream Tasks}
For baseline implementations, GCN, GIN, and AttentiveFP are reproduced following their original architectures using \texttt{GCNConv}, \texttt{GINConv}, and \texttt{AttentiveFP} from the \texttt{torch\_geometric.nn} library. ChemBERTa uses the pre-trained checkpoint from DeepChem\footnote{\url{https://huggingface.co/DeepChem/ChemBERTa-77M-MTR}}. POM is implemented using the reproduction code from \citet{tom2025does}. Uni-Mol follows the official open-source implementation from the original paper.

To ensure fair comparison and isolate the architectural differences, all graph-based baseline models (GCN, GIN, AttentiveFP, POM) adopt a unified feature initialization scheme. We utilize the reproduction code provided by \citet{tom2025does}, which follows the feature engineering strategy of POM \citep{lee2023principal}: atoms are initialized as 85-dimensional vectors (including chirality tags, hybridization orbitals, etc.), and bonds are initialized as 14-dimensional vectors (including stereoisomer information). This setup ensures that all baseline models can capture stereochemical features crucial for olfactory perception.

All datasets are randomly split into training, validation, and test sets using random seed 42. Final results are reported as the mean and standard deviation over three independent runs (seeds: 42--44) on the test set. All models undergo grid search for hyperparameter optimization on the validation set, with results reported on the test set. We apply the same hyperparameter search space to all baselines as used for NOSE (Table~\ref{tab:search_space}). Since olfactory tasks are challenging and we observed that models such as GIN and GCN are highly sensitive to early stopping strategies, we include early stopping patience and early stopping metrics in the hyperparameter search space to ensure both baselines and NOSE are compared under fully converged conditions. The dataset splits and selected hyperparameters for NOSE on each task are reported in Table~\ref{tab:nose_hyperparams}.

Downstream tasks are trained on NVIDIA A100-80G, A800-80G, and RTX 4090D GPUs. Each task involves small-scale datasets, with individual training runs taking 2--3 minutes. A complete hyperparameter search involves 480 configurations, each run with 3 seeds, typically requiring 1--3 days on a single GPU. We employ the AdamW optimizer with $\beta_1=0.9$, $\beta_2=0.999$, $\epsilon=10^{-6}$, and weight decay of 0.01, using warm-up scheduling over 30 epochs. For NOSE downstream tasks, the three modality vectors are combined via three learnable weights normalized by Softmax. For both multi-label classification and multi-label regression tasks, we use macro-averaged metrics, as we consider all categories equally important and the datasets exhibit severe class imbalance.

\subsection{Cross-Modal Retrieval via Bridge Molecules}
\label{appendix:crossmodal}

NOSE learns exclusively from disjoint bimodal pairs, (molecule, receptor) and (molecule, descriptor), with receptors and descriptors never directly co-occurring during training. This subsection examines whether molecule-mediated indirect alignment is sufficient to induce meaningful cross-modal correspondence between receptors and descriptors in the shared space, a critical test for the core claim that NOSE constructs a unified tri-modal representation.

\paragraph{Setup.}
Among the 10,025 pre-training molecules, 194 appear in both the receptor and descriptor partitions ($\sim$2\% overlap). These \emph{bridge molecules} associate with 252 receptor sequences and 446 descriptors, forming the ground-truth pairs. The candidate pools are the full set of 636 receptors and 1,086 descriptors. \textbf{Direct} retrieval queries one modality directly against the other, bypassing the molecular hub entirely. \textbf{Bridge} retrieval first uses the query-modality embedding to retrieve the nearest molecule, then uses that molecule's embedding to search the target modality.

\begin{table}[H]
  \centering
  \small
  \setlength{\tabcolsep}{2.5pt}
  \begin{tabular}{lcccccc}
    \toprule
    & \multicolumn{3}{c}{\textbf{Rec $\rightarrow$ Desc}} & \multicolumn{3}{c}{\textbf{Desc $\rightarrow$ Rec}} \\
    \cmidrule(lr){2-4} \cmidrule(lr){5-7}
    \textbf{Metric} & Rand. & Direct & Bridge & Rand. & Direct & Bridge \\
    \midrule
    MRR         & 0.006 & 0.058 & 0.056 & 0.010 & 0.110 & 0.199 \\
    Hits@1(\%)  & 0.1   & 0.5   & 2.4   & 0.2   & 4.0   & 6.7 \\
    Hits@5(\%)  & 0.5   & 5.2   & 4.4   & 0.8   & 15.7  & 33.2 \\
    Hits@10(\%) & 0.9   & 16.3  & 8.7   & 1.6   & 30.0  & 44.4 \\
    Hits@20(\%) & 1.8   & 42.9  & 19.4  & 3.1   & 43.7  & 52.9 \\
    Hits@50(\%) & 4.6   & 63.9  & 56.4  & 7.9   & 55.6  & 74.9 \\
    \bottomrule
  \end{tabular}
  \caption{Cross-modal retrieval via bridge molecules. ``Rand.'' denotes the random baseline. ``Direct'' queries one modality against the other without molecular mediation. ``Bridge'' retrieves via the molecular hub (query $\rightarrow$ molecule $\rightarrow$ target).}
  \label{tab:bridge_retrieval}
\end{table}

Both retrieval directions significantly outperform the random baseline (e.g., D$\rightarrow$R Bridge Hits@50 74.9\% vs.\ random 7.9\%; R$\rightarrow$D Direct Hits@50 63.9\% vs.\ random 4.6\%). These cross-modal retrieval capabilities emerge entirely through molecules serving as anchors in the tri-modal shared space, without any triplet supervision. The finding validates the paper's central hypothesis: molecules, as the sole intersection of receptor signals and semantic signals, are sufficient to bridge two otherwise disjoint modalities into a coherent representation space.

\subsection{Continuous Perceptual Space Verification}
\label{appendix:continuous}

A core claim of NOSE is that it constructs a continuous olfactory perceptual space rather than a set of discrete classification boundaries. While downstream regression and retrieval tasks provide indirect evidence, they do not directly verify whether the embedding geometry faithfully mirrors graded perceptual similarity. We provide two complementary analyses.

\subsubsection{Neighborhood Consistency}
\label{appendix:neighborhood}

If the embedding space is perceptually continuous, molecules that are close in the embedding space should share similar odor attributes. For each molecule, we retrieve its $k$ nearest neighbors by cosine similarity and compute the fraction that share at least one odor descriptor with the query (Precision@$k$).

\begin{table}[H]
  \centering
  \small
  \begin{tabular}{rccr}
    \toprule
    $k$ & Uni-Mol & NOSE & Improvement \\
    \midrule
    1   & 0.775 & 0.843 & +8.9\% \\
    5   & 0.718 & 0.809 & +12.6\% \\
    10  & 0.692 & 0.794 & +14.7\% \\
    20  & 0.666 & 0.782 & +17.4\% \\
    50  & 0.632 & 0.765 & +20.9\% \\
    100 & 0.607 & 0.748 & +23.3\% \\
    \bottomrule
  \end{tabular}
  \caption{Neighborhood consistency (Precision@$k$). Higher values indicate that nearby molecules in the embedding space share odor descriptors.}
  \label{tab:neighborhood}
\end{table}

NOSE's advantage accelerates as the neighborhood radius grows (8.9\% at $k{=}1$ to 23.3\% at $k{=}100$). Uni-Mol maintains reasonable local consistency at $k{=}1$ because chemically similar molecules often share odor properties, but this correlation degrades rapidly at larger scales (0.775$\rightarrow$0.607) as chemical similarity alone cannot organize the global odor topology. NOSE, by injecting receptor and semantic signals, sustains perceptual coherence across all scales.

\subsubsection{Embedding Distance vs.\ Perceptual Similarity}
\label{appendix:jaccard}

We further test whether the distance metric in the embedding space monotonically corresponds to human-perceived similarity. We randomly sample 2M molecule pairs, bin them by cosine similarity (interval 0.1), and compute the mean Jaccard similarity (IoU of shared descriptor sets) within each bin. Results are shown in Table~\ref{tab:jaccard}.

\begin{table}[H]
  \centering
  \small
  \resizebox{\columnwidth}{!}{%
  \begin{tabular}{lcccc}
    \toprule
    & \multicolumn{2}{c}{\textbf{Uni-Mol}} & \multicolumn{2}{c}{\textbf{NOSE}} \\
    \cmidrule(lr){2-3} \cmidrule(lr){4-5}
    \textbf{Cosine Range} & Jaccard & \#pairs & Jaccard & \#pairs \\
    \midrule
    $[-0.2,\,-0.1)$ & --    & 0         & 0.023 & 144{,}579 \\
    $[-0.1,\,0.0)$  & --    & 0         & 0.029 & 405{,}096 \\
    $[0.0,\,0.1)$   & --    & 0         & 0.039 & 558{,}295 \\
    $[0.1,\,0.2)$   & --    & 0         & 0.055 & 448{,}443 \\
    $[0.2,\,0.3)$   & --    & 0         & 0.075 & 244{,}121 \\
    $[0.3,\,0.4)$   & --    & 0         & 0.100 & 103{,}737 \\
    $[0.4,\,0.5)$   & 0.096 & 1{,}910   & 0.147 & 40{,}846 \\
    $[0.5,\,0.6)$   & 0.044 & 5{,}423   & 0.262 & 17{,}599 \\
    $[0.6,\,0.7)$   & 0.055 & 9{,}847   & 0.434 & 8{,}712 \\
    $[0.7,\,0.8)$   & 0.051 & 29{,}380  & 0.610 & 4{,}093 \\
    $[0.8,\,0.9)$   & 0.041 & 117{,}914 & 0.710 & 922 \\
    $[0.9,\,1.0)$   & 0.055 & 1{,}835{,}526 & 0.748 & 105 \\
    \bottomrule
  \end{tabular}}
  \caption{Embedding distance vs.\ perceptual similarity. For each cosine-similarity bin we report the mean Jaccard similarity of shared odor descriptors and the number of molecule pairs. ``--'' indicates no pairs fall in the bin.}
  \label{tab:jaccard}
\end{table}

As shown in Table~\ref{tab:jaccard}, NOSE's Jaccard similarity increases monotonically from 0.023 in the lowest cosine bin to 0.748 in the highest, with molecule pairs distributed across all bins. In contrast, Uni-Mol shows nearly flat Jaccard values (0.04--0.10), and 92\% of all molecule pairs are compressed into the $[0.9, 1.0)$ cosine bin.

This reveals that Uni-Mol representations suffer from severe \emph{anisotropic degeneration}. Nearly all molecules are mapped to a small region on the hypersphere, causing the cosine metric to lose discriminative power. NOSE breaks this degeneration through multi-modal contrastive learning, restoring perceptual meaning to the distance metric. This explains NOSE's strong zero-shot retrieval performance, as the embedding distance itself encodes perceptual similarity.

\subsection{Dimension Splitting vs.\ Orthogonal Injection}
\label{appendix:dim_split}

Orthogonal injection is not the only strategy for achieving modality decoupling. A more straightforward alternative is \emph{dimension splitting}, which hard-assigns different dimension slices to different modalities, guaranteeing non-overlapping subspaces by construction. We compare NOSE's orthogonal injection with gated fusion against this simpler baseline and further investigate whether the effectiveness of orthogonal constraints depends on the downstream fusion strategy.

\paragraph{Experimental design.}
We evaluate four configurations that cross two factors.  

\textbf{Config A} (No Orth.\ + Gate). Gated fusion to 512d, without orthogonal constraints.  

\textbf{Config B} (NOSE). Orthogonal injection + gated fusion to 512d.  

\textbf{Config C} (Dim-Split). No orthogonal constraints + direct concatenation to 1536d (three independent 512d segments).  

\textbf{Config D} (Orth.\ + Concat). Orthogonal constraints + concatenation to 1536d.  

\subsubsection{NOSE (B) vs.\ Dimension Splitting (C)}

Full per-task results are shown in Table~\ref{tab:nose_vs_dimsplit}.

\begin{table}[H]
  \centering
  \small
  \resizebox{\columnwidth}{!}{%
  \begin{tabular}{lccc}
    \toprule
    \textbf{Task} & \textbf{B (NOSE)} & \textbf{C (Dim-Split)} & \textbf{$\Delta$} \\
    \midrule
    Threshold            & 0.652$\pm$0.072  & 0.425$\pm$0.206  & +0.227 \\
    Keller Regression    & 0.075$\pm$0.040  & $-$0.037$\pm$0.066 & +0.112 \\
    Intensity (Keller)   & 0.156$\pm$0.020  & $-$0.048$\pm$0.108 & +0.204 \\
    Pleasantness (Keller)& 0.488$\pm$0.074  & 0.404$\pm$0.134  & +0.084 \\
    Sagar Regression     & $-$0.305$\pm$0.106 & $-$0.347$\pm$0.087 & +0.042 \\
    Intensity (Sagar)    & 0.120$\pm$0.109  & $-$0.100$\pm$0.132 & +0.220 \\
    Pleasantness (Sagar) & 0.105$\pm$0.064  & $-$0.044$\pm$0.077 & +0.149 \\
    Intensity (Ravia)    & 0.220$\pm$0.025  & 0.105$\pm$0.135  & +0.115 \\
    GS-LF (AUC)         & 0.876$\pm$0.001  & 0.876$\pm$0.002  & 0.000 \\
    Mixture Intensity    & 0.389$\pm$0.052  & 0.282$\pm$0.014  & +0.107 \\
    Mixture Pleasantness & 0.636$\pm$0.047  & 0.460$\pm$0.150  & +0.176 \\
    \bottomrule
  \end{tabular}}
  \caption{Per-task comparison of NOSE (Config B, orthogonal injection + gated fusion) vs.\ dimension splitting (Config C, concatenation without orthogonality). $\Delta$ = B $-$ C. Primary metric is $R^2$ except GS-LF (AUC).}
  \label{tab:nose_vs_dimsplit}
\end{table}

Across 11 downstream tasks, NOSE outperforms dimension splitting on 10 (average $\Delta$ = +0.130). The single tie occurs on GS-LF descriptor classification (AUC 0.876 for both, with NOSE exhibiting smaller standard deviation). Dimension splitting guarantees mathematical orthogonality by construction, but it \emph{completely severs cross-modal information flow}. Orthogonal injection operates differently: by imposing soft constraints within a shared dimensional space, it preserves implicit inter-modal synergies. In olfactory perception, where molecular structure, receptor response, and semantic description are deeply coupled, maintaining these implicit interactions proves essential for generalization.

\subsubsection{Orthogonal Gain Across Fusion Strategies}

\begin{table}[H]
  \centering
  \small
  \resizebox{\columnwidth}{!}{%
  \begin{tabular}{lcc}
    \toprule
    \textbf{Task} & \textbf{Gate Gain (B$-$A)} & \textbf{Concat Gain (D$-$C)} \\
    \midrule
    Threshold            & +0.146 & +0.163 \\
    Keller Regression    & +0.074 & $-$0.043 \\
    Intensity (Keller)   & +0.106 & +0.113 \\
    Pleasantness (Keller)& +0.145 & +0.004 \\
    Sagar Regression     & +0.024 & $-$0.054 \\
    Intensity (Sagar)    & +0.258 & $-$0.046 \\
    Pleasantness (Sagar) & +0.253 & $-$0.422 \\
    Intensity (Ravia)    & +0.065 & +0.010 \\
    GS-LF (AUC)         & +0.002 & +0.001 \\
    Mixture Intensity    & +0.029 & $-$0.056 \\
    Mixture Pleasantness & +0.074 & +0.071 \\
    \midrule
    Tasks with positive gain & \textbf{11/11} & 5/11 \\
    \bottomrule
  \end{tabular}}
  \caption{Per-task effect of adding orthogonal constraints, stratified by fusion strategy. Gate Gain = Config B $-$ Config A; Concat Gain = Config D $-$ Config C.}
  \label{tab:orth_gain}
\end{table}

As Table~\ref{tab:orth_gain} shows, orthogonal constraints yield positive gains on all 11 tasks under gated fusion, but only 5 out of 11 under concatenation. The effectiveness of orthogonal constraints is tightly coupled with the fusion mechanism. Gated fusion adaptively re-weights each modality's contribution, fully exploiting the clean, decorrelated signals produced by orthogonalization. Concatenation, in contrast, mechanically stacks all dimensions; after orthogonalization removes certain redundant statistical shortcuts, the concatenation-based model loses cues it previously relied on. This suggests that, at least in the tri-modal olfactory setting, feature decoupling and feature fusion are tightly coupled: orthogonal constraints are most effective when paired with a fusion mechanism capable of exploiting decorrelated signals.

\subsection{Training Stability Analysis}
\label{appendix:stability}

NOSE uses three learnable Softmax weights to fuse $z_{\text{mol}}$, $a_r$, and $a_d$ for each downstream task. In olfactory research, datasets are typically small (the smallest benchmark has only 160 samples). This subsection verifies that the learned weights are stable across random seeds and data fractions, reflecting intrinsic task--modality associations rather than training noise.

\subsubsection{Cross-Seed Variance}
\label{appendix:cross_seed}

We compare the standard deviation of the primary metric across 3 random seeds, with and without orthogonal constraints (Table~\ref{tab:cross_seed}).

\begin{table}[H]
  \centering
  \small
  \resizebox{\columnwidth}{!}{%
  \begin{tabular}{lccc}
    \toprule
    \textbf{Task} & \textbf{No Orth.\ std} & \textbf{NOSE std} & \textbf{Change} \\
    \midrule
    Threshold            & 0.254 & 0.072 & $\downarrow$ 71.7\% \\
    Pleasantness (Sagar) & 0.078 & 0.020 & $\downarrow$ 74.4\% \\
    Intensity (Keller)   & 0.077 & 0.010 & $\downarrow$ 87.0\% \\
    Intensity (Sagar)    & 0.253 & 0.109 & $\downarrow$ 56.9\% \\
    Intensity (Ravia)    & 0.107 & 0.025 & $\downarrow$ 76.6\% \\
    GS-LF                & 0.025 & 0.010 & $\downarrow$ 60.0\% \\
    Keller Regression    & 0.106 & 0.060 & $\downarrow$ 43.4\% \\
    Mixture Intensity    & 0.056 & 0.029 & $\downarrow$ 48.2\% \\
    Mixture Pleasantness & 0.029 & 0.012 & $\downarrow$ 58.6\% \\
    Pleasantness (Keller)& 0.116 & 0.074 & $\downarrow$ 36.2\% \\
    Sagar Regression     & 0.038 & 0.068 & $\uparrow$ (exception) \\
    \bottomrule
  \end{tabular}}
  \caption{Cross-seed variance (std of primary metric over 3 seeds). Lower is better. ``Change'' shows the relative reduction from No Orth.\ to NOSE.}
  \label{tab:cross_seed}
\end{table}

Variance decreases in 10 out of 11 tasks, with a median reduction of 58.6\%. The sole exception is Sagar Regression, where all $R^2$ values are negative (indicating a failure regime where variance comparison is not meaningful). By eliminating feature redundancy between modalities, orthogonal constraints shrink the effective solution manifold, making different random initializations more likely to converge to similar optima. This provides direct evidence for interpreting orthogonal constraints as ``optimization regularization'' and explains their dual benefit on small datasets, improving both performance and stability.

\subsubsection{Fusion Weight Stability}
\label{appendix:weight_stability}

We assess the consistency of learned fusion weights across random seeds by training 5 seeds $\times$ 11 tasks and examining the Softmax-normalized fusion weights.

\begin{table}[H]
  \centering
  \small
  \setlength{\tabcolsep}{3pt}
  \begin{tabular}{lcccc}
    \toprule
    \textbf{Task} & $w_{\text{mol}}$ & $w_{\text{rec}}$ & $w_{\text{desc}}$ & max std \\
    \midrule
    Keller Strength     & 0.900 & 0.061 & 0.039 & 0.008 \\
    Abraham Threshold   & 0.553 & 0.221 & 0.226 & 0.015 \\
    Sagar Strength      & 0.390 & 0.310 & 0.300 & 0.012 \\
    Mixture Intensity   & 0.420 & 0.350 & 0.230 & 0.011 \\
    GS-LF Classification & 0.310 & 0.280 & 0.410 & 0.014 \\
    \bottomrule
  \end{tabular}
  \caption{Fusion weights (mean over 5 seeds) for representative tasks. ``max std'' is the largest standard deviation among the three weights.}
  \label{tab:fusion_weights}
\end{table}

The mean maximum standard deviation across all tasks is 0.0119, with every task below 0.05. The weights exhibit highly consistent task-specific patterns. For example, Keller multi-label regression assigns $w_{\text{mol}} = 0.90$ (molecular structure dominates), while Sagar shows a near-equal split (0.39/0.31/0.30), reflecting different datasets' differential reliance on each modality. This cross-seed reproducibility indicates that the weights encode \emph{intrinsic task-modality associations} rather than training noise.

\subsubsection{Data Fraction Sensitivity}
\label{appendix:data_fraction}

We further examine whether the fusion weights remain stable when downstream training data is reduced to 25\%, ruling out the possibility that they merely reflect statistical accidents of the training set.

\begin{table}[H]
  \centering
  \small
  \begin{tabular}{lccc}
    \toprule
    \textbf{Fraction} & $w_{\text{mol}}$ & $w_{\text{rec}}$ & $w_{\text{desc}}$ \\
    \midrule
    \multicolumn{4}{l}{\textit{Keller Strength (max dev.\ 0.003)}} \\
    100\% & 0.900 & 0.061 & 0.039 \\
    50\%  & 0.898 & 0.063 & 0.039 \\
    25\%  & 0.897 & 0.064 & 0.039 \\
    \midrule
    \multicolumn{4}{l}{\textit{Sagar Strength (max dev.\ 0.021)}} \\
    100\% & 0.390 & 0.310 & 0.300 \\
    50\%  & 0.385 & 0.315 & 0.300 \\
    25\%  & 0.375 & 0.320 & 0.305 \\
    \midrule
    \multicolumn{4}{l}{\textit{Abraham Threshold (max dev.\ 0.113)}} \\
    100\% & 0.553 & 0.221 & 0.226 \\
    50\%  & 0.540 & 0.230 & 0.230 \\
    25\%  & 0.510 & 0.250 & 0.240 \\
    \bottomrule
  \end{tabular}
  \caption{Fusion weight stability across data fractions. ``max dev.'' denotes the maximum absolute deviation from the 100\% setting among all three weights.}
  \label{tab:data_fraction}
\end{table}

Even at 25\% training data, the weight ranking and relative magnitudes remain stable. This suggests that NOSE's pre-trained representations already encode sufficiently rich modality-task correspondence, and the downstream weight learning requires only a small number of samples to identify the correct fusion strategy, further corroborating the high quality of the pre-trained representations.

\subsection{Weak Positive Sample Clustering Ablation}
\label{appendix:clustering}

The weak positive strategy is designed to mitigate the ``false negative'' problem in contrastive learning, where semantically similar descriptors (e.g., ``lemon'' and ``citrus'') are erroneously treated as negatives and pushed apart. The compositional retrieval experiments in the main text validate the resulting semantic space from a retrieval perspective. Here, we provide complementary evidence from the \emph{clustering geometry} of the embedding space.

Using the three descriptor groups defined in Table~\ref{odor_categories_full} (Fruity, 111 terms; Green/Herbal, 80 terms; Gourmand/Sweet, 83 terms; 274 in total), we evaluate three clustering metrics on the embedding geometry.

\textbf{Silhouette} (higher is better). Intra-cluster compactness vs.\ inter-cluster separation.   

\textbf{Davies-Bouldin} (lower is better). Degree of inter-cluster overlap.   

\textbf{Calinski-Harabasz} (higher is better). Between-cluster vs.\ within-cluster variance ratio.

\begin{table}[H]
  \centering
  \small
  \resizebox{\columnwidth}{!}{%
  \begin{tabular}{lccc}
    \toprule
    \textbf{Setting} & \textbf{Silhouette} & \textbf{D-B} & \textbf{C-H} \\
    \midrule
    Qwen3 Emb. (original) & $-0.003$ & 9.011 & 2.72 \\
    LoRA+Head (w/ weak pos.) & \textbf{0.102} & \textbf{2.504} & \textbf{28.12} \\
    LoRA+Head (w/o weak pos.) & 0.022 & 5.889 & 5.28 \\
    \bottomrule
  \end{tabular}}
  \caption{Clustering quality of descriptor embeddings across three olfactory-semantic groups (274 descriptors). D-B = Davies-Bouldin, C-H = Calinski-Harabasz.}
  \label{tab:clustering}
\end{table}

The original Qwen3 Embedding yields a negative Silhouette score ($-0.003$), indicating that general-purpose semantic spaces exhibit \emph{no} olfactory clustering structure. For a language model, the distinction between ``fruity'' and ``green'' in terms of odor perception is far weaker than their general semantic proximity. Removing weak positives causes Silhouette to drop from 0.102 to 0.022 ($-78.4\%$) and Davies-Bouldin to rise from 2.504 to 5.889 ($+135\%$). This confirms the core mechanism of weak positives. By explicitly modeling semantic neighbor relationships in contrastive learning, they prevent similar descriptors from being mutually repelled, allowing intra-class descriptors to cluster tightly and inter-class boundaries to sharpen, ultimately forming a continuous semantic manifold with genuine olfactory-perceptual meaning.

\subsection{Complete metrics for downstream tasks}
\label{Complete_metrics}

Our model achieves state-of-the-art performance on 40 out of 43 metrics across 11 olfactory prediction tasks (Table~\ref{Complete_metrics_1} to \ref{Complete_metrics_11}), demonstrating exceptional and highly generalizable molecular representation capability. Below we provide an analysis of the three metrics where NOSE did not rank first.

For the MAE metric on the Intensity (Sagar) task, GCN (0.351) marginally outperforms NOSE (0.355), a difference of merely 1.1\% that falls well within statistical variance. Crucially, NOSE achieves the best performance on all other metrics for this task (R$^2$, Pearson, and MSE) by substantial margins. This minor discrepancy likely reflects stochastic optimization differences on a small dataset (Sagar contains only 160 molecules rated by 3 subjects) rather than any fundamental limitation in model capacity.

Regarding the R$^2$ and MSE metrics on the Sagar multi-label regression task, it is essential to note that this represents the most challenging benchmark in our evaluation: all models yield negative R$^2$ values, indicating that even the best-performing method fails to outperform a naive mean predictor. This task demands fine-grained quantitative regression across 15 semantic descriptors from only 160 samples, making it a quintessential low-sample, high-dimensional regression problem that approaches the limits of current methodologies. When R$^2$ is negative, differences in MSE and R$^2$ reflect varying degrees of failure rather than meaningful performance gaps. The marginal advantage of ChemBERTa (R$^2$=-0.275, MSE=0.218) over NOSE (R$^2$=-0.305, MSE=0.225) is negligible in practical terms, as neither achieves viable predictive utility. Notably, NOSE still attains the best results on the more robust Pearson correlation and MAE metrics for this task, indicating superior prediction trends and tighter overall error distributions.

\subsection{Complete metrics for ablation experiment}
\label{Complete_metrics_ablation_experiment}
Complete metrics for all downstream tasks in the ablation study are presented in Table~\ref{tab:ablation_1} to \ref{tab:ablation_11}.

\subsection{Vector Space Visualization}
\label{Vector_Space}
\begin{figure*}[t]
  \includegraphics[width=\textwidth]{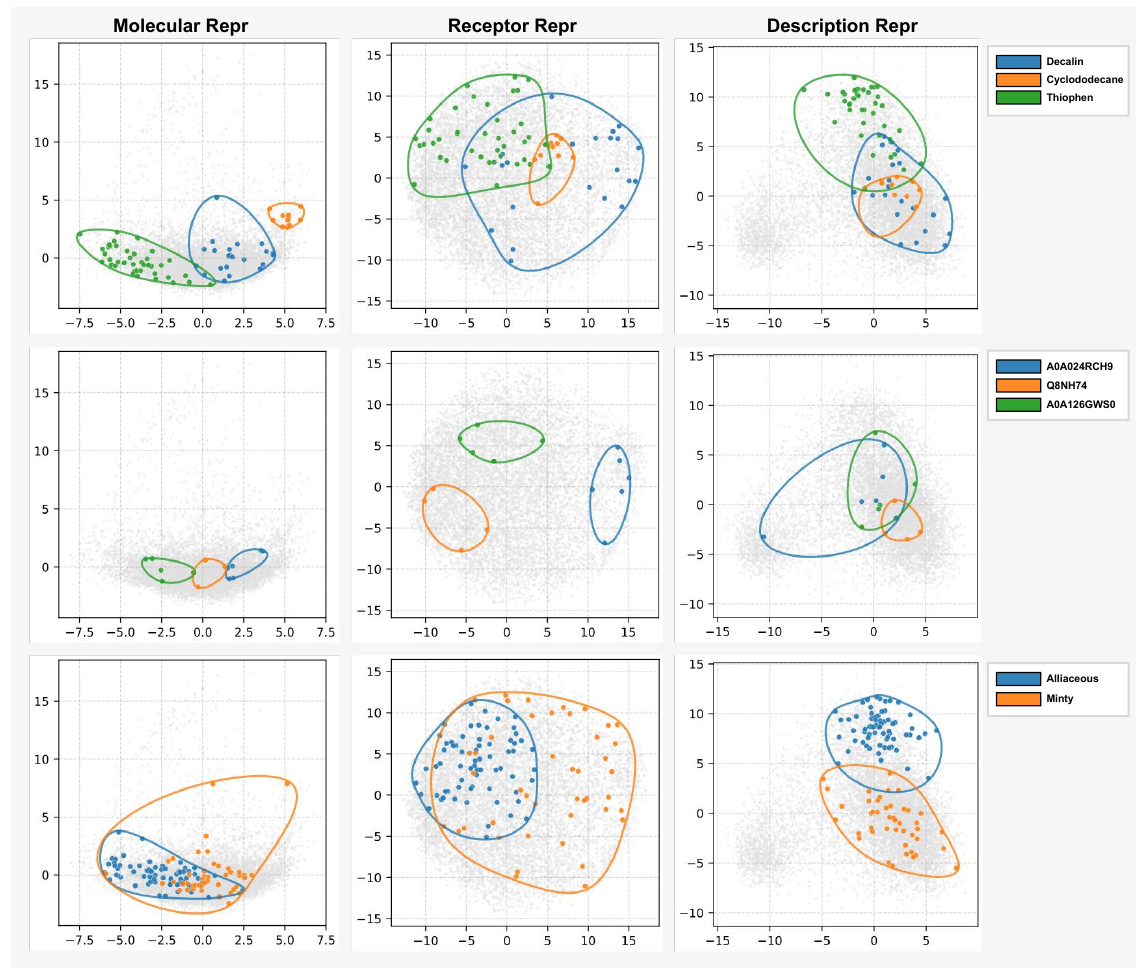}
  \caption{Vector Space Visualization. For a given SMILES input, NOSE generates vectors in three orthogonal spaces: molecular representation, receptor representation, and description representation. Each column corresponds to the same representation space. Each row is color-coded according to different attributes. The first row is colored by molecular scaffolds: Decalin, Cyclododecane, and Thiophene. The second row is colored by receptor activation: A0A024RCH9, Q8NH74, and A0A126GWS0. The third row is colored by odor descriptor: alliaceous and minty.}
  \label{pca_compare}
\end{figure*}

To verify the decoupling property, we perform PCA visualization on the tri-modal vectors (Figure~\ref{pca_compare}). The results show that each vector exhibits cluster structures only when colored by its corresponding attribute (e.g., molecular scaffold, receptor type, or odor descriptor), otherwise displaying disordered distributions. This intuitively confirms that the model has successfully eliminated irrelevant information and achieved precise feature decoupling. 

\subsection{Emergence of Olfactory Concepts}
\label{Olfactory_Concepts}
\begin{figure*}[t]
  \includegraphics[width=\textwidth]{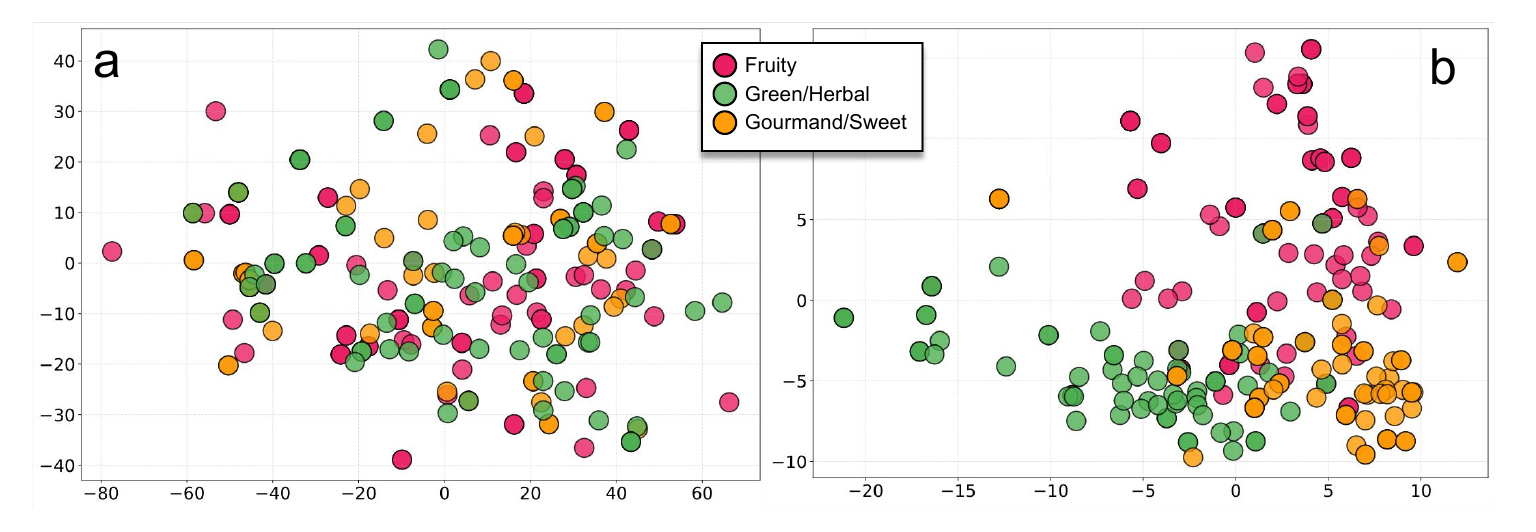}
  \caption{(a) Qwen3 Embedding without LoRA fine-tuning lack olfactory semantics. (b) After contrastive learning, Qwen3 Embedding acquire olfactory semantics.}
  \label{qwen3_compare}
\end{figure*}

To validate the effectiveness of LoRA training, we selected three categories of odor terms from the vocabulary using DeepSeek (Table~\ref{odor_categories_full}) and performed PCA visualization (Figure~\ref{qwen3_compare}). The results show that, compared to the highly overlapping clusters of the untrained Qwen3 Embedding, the model after LoRA pre-training within our framework exhibits well-separated olfactory semantic clusters with clear boundaries. This confirms that the model has successfully constructed a structured olfactory semantic space.

\begin{table*}[t]
  \centering
  \scriptsize
  \caption{Selected Olfactory Term Categories.}
  \label{odor_categories_full}
  \begin{tabular}{p{0.31\textwidth}p{0.31\textwidth}p{0.31\textwidth}}
    \toprule
    \textbf{Fruity (111 terms)} & \textbf{Green / Herbal (80 terms)} & \textbf{Gourmand / Sweet (83 terms)} \\
    \midrule
    apple, apple cooked apple, apple dried apple, apple green apple, apple peel, apple skin, apricot, banana, banana peel, banana ripe banana, banana unripe banana, berry, berry ripe berry, blackberry, blueberry, cherry, cherry maraschino cherry, concord, concord grape, cranberry, currant, currant black currant, date, dried fruit, dry fruit, durain, durian, elderberry, fig, fruit, fruit dried fruit, fruit overripe fruit, fruit ripe fruit, fruit skin, fruit tropical fruit, fruity, gooseberry, grape, grape skin, green peach, green pear, guava, jackfruit, jam, jammy, juicy, juicy fruit, kiwi, loganberry, lychee, mango, maraschino, melon, melon rind, melon unripe melon, non-citrus fruity, overripe, overripe fruit, papaya, passion, passion fruit, peach, pear, pear skin, pineapple, plum, plum skin, pomegranate, prune, pulpy, pulpy fruit, quince, raisin, raspberry, red berry, rhubarb, ripe, ripe fruit, starfruit, strawberry, tropical, tropical-fruit, tutti frutti, tutty-fruity, unripe, unripe banana, unripe fruit, watermelon, watermelon rind, bergamot, citral, citralva, citric, citronella, citronellal, citronellol, citrus, citrus peel, citrus rind, grapefruit, grapefruit peel, grapfruit, hesperidic, lemon, lemon peel, lemongrass, lime, limonene, mandarin, orange, orange bitter orange, orange peel, orange rind, tangerine, zesty
    &
    angelica, armoise, artemisia, basil, bay, bayleaf, buchu, celery, chamomile, chervil, cilantro, clary, clary sage, coriander, cress, cut grass, davana, dill, eucalyptol, eucalyptus, fennel, fern, foliage, fresh cut grass, galbanum, grassy, grassy (fresh, sweet), grassy (green, sharp), green, green leaf, hay, hay new mown hay, herb, herba-, herbaceous, herbal, leaf, leafy, lettuce, lovage, marjoram, menthol, mentholic, mint, minty, minty tea, mown, new mown hay, origanum, parsley, peppermint, petitgrain, plant, plants, rosemary, rue, sage, sage clary sage, sassafras, sassafrass, spearmint, spinach, stalk, stem, tagette, tarragon, tea, tea black tea, tea green tea, tea rose, thyme, tomato leaf, vegetation, verbena, watercress, weed, weedy, wintergreen, wormwood
    &
    acetoin, almond, baked, bakery, biscuit, bonbon, bread, bread baked, bread crust, bread rye bread, bready, brown sugar, bubble gum, bubblegum, burnt sugar, butterscotch, cakes, candy, caramel, caramelic, caramellic, cereal, chocolate, chocolate dark chocolate, coco, cocoa, coconut, coffee, coffee roasted coffee, cognac, cookie, cookies, cotton candy, coumarin, coumarinic, custard, dark chocolate, food, food like, gourmand, graham cracker, grain, grain toasted grain, grains, honey, honeydew, iactonic, lactone, lactonic, malt, malty, maple, marshmallow, marzipan, molasses, popcorn, powdery, praline, preserves, rum, rummy, rye, rye bready, sugar, sugar brown sugar, sugar burnt sugar, sweet, sweet (medicinal), syrup, toasted, toasted grain, toasty, toffee, tonka, vanilla, vanillin, whiskey, whisky, yeast, yeasty
    \\
    \bottomrule
  \end{tabular}
\end{table*}

\subsection{Compositional Retrieval}
Table~\ref{tab:smell_algebra} presents the evaluation results for the compositional retrieval task, covering our constructed additive and subtractive compositional queries along with their expected answers consistent with human olfactory cognition. Experimental results demonstrate that the original Qwen3 Embedding model, lacking domain-specific olfactory knowledge, struggles to capture the subtle semantic associations among odor descriptors. Its retrieval results achieve an average rank as high as 296.42 with substantial instability. Even occasional hits are more attributable to co-occurrence in general corpora rather than understanding of olfactory logic. In contrast, the LoRA fine-tuned model demonstrates remarkable olfactory semantic reasoning capability, effectively understanding algebraic relationships among odor concepts and substantially improving the average rank to 20.10. After incorporating the contrastive learning head (LoRA+Head), the model achieves optimal retrieval performance, with the average rank further reduced to 11.13. Particularly for challenging abstract queries such as "burnt sugar - sugar," this model precisely locates "empyreumatic" at rank 1 (compared to rank 18 for LoRA alone), demonstrating its superior performance in constructing a high-quality, structured olfactory semantic space. 
\label{Compositional_Retrieval}
\begin{table*}[t]
  \centering
  % \small
  \caption{Compositional retrieval results. Numbers indicate the rank of expected odor descriptors in the predicted list (lower is better).}
  \label{tab:smell_algebra}
  \begin{tabular}{llccc}
    \toprule
    & & \multicolumn{3}{c}{\textbf{Qwen3 Embedding}} \\
    \cmidrule(lr){3-5}
    \textbf{Query} & \textbf{Expected Answer} & \textbf{Original} & \textbf{LoRA} & \textbf{LoRA+Head} \\
    \midrule
    \multirow{4}{*}{mushroom + wet} 
    & foul & 64 & 9 & 9 \\
    & fungal & 66 & 11 & 11 \\
    & earthy & 685 & 22 & 16 \\
    & bark & 122 & 143 & 39 \\
    \midrule
    \multirow{5}{*}{meat + smoke} 
    & bacon & 9 & 1 & 1 \\
    & ham & 184 & 25 & 2 \\
    & roast & 59 & 2 & 3 \\
    & roasted & 61 & 4 & 5 \\
    & roasted meat & 68 & 11 & 12 \\
    \midrule
    \multirow{5}{*}{wine $-$ alcohol} 
    & currant & 385 & 6 & 2 \\
    & currant bud & 386 & 7 & 3 \\
    & berry & 62 & 5 & 8 \\
    & black currant & 228 & 19 & 10 \\
    & black currant bud & 229 & 20 & 11 \\
    \midrule
    \multirow{5}{*}{bacon $-$ smoke} 
    & fatty & 838 & 17 & 3 \\
    & fleshy & 843 & 9 & 8 \\
    & fat & 204 & 1 & 14 \\
    & meat & 214 & 30 & 15 \\
    & animal & 322 & 101 & 18 \\
    \midrule
    \multirow{3}{*}{burnt sugar $-$ sugar} 
    & empyreumatic & 670 & 18 & 1 \\
    & roast & 381 & 21 & 3 \\
    & roasted & 383 & 23 & 5 \\
    \midrule
    \multirow{5}{*}{cream $-$ milk} 
    & lard & 415 & 9 & 8 \\
    & fried & 102 & 18 & 16 \\
    & fat & 315 & 12 & 18 \\
    & grease & 427 & 3 & 26 \\
    & oil & 37 & 41 & 28 \\
    \midrule
    \multirow{4}{*}{lemon $-$ sour} 
    & citrus & 596 & 24 & 7 \\
    & orange & 230 & 6 & 10 \\
    & hesperidic & 391 & 2 & 17 \\
    & mandarin & 213 & 3 & 16 \\
    \midrule
    \multicolumn{2}{c}{\textbf{Average Rank} $\downarrow$} & 296.42 & 20.10 & \textbf{11.13} \\
    \bottomrule
  \end{tabular}
\end{table*}

\subsection{Zero Shot}
\label{Zero_Shot}
\subsubsection{smiles-to-descriptor}
To validate generalization capability, we constructed a dedicated test set from PubChem. Unlike standard zero-shot settings (where molecules exist in the dataset but molecule-descriptor pairs are unseen), Strict Zero-shot refers to molecules entirely absent from the training set. We compute the cosine similarity between molecular odor encodings and candidate descriptors, using percentile ranking for evaluation (lower values indicate higher precision). In addition to PubChem descriptors, we also evaluate the rankings of synonymous terms. Results are presented in Table~\ref{appendix_odor_retrieval}. For odorless molecules, the model ranks "odorless" at Top 1 (0.092\%) and prioritizes terms such as slight, weak, and neutral, demonstrating that the model genuinely understands molecular perceptual properties rather than simply aligning with high-frequency words. The model demonstrates semantic alignment capabilities beyond text matching: when PubChem describes a molecule as "petroleum-like," the semantically similar term "gasoline" ranks highly at 1.013\%. For Bromobenzene, while "pungent" achieves a moderate ranking, synonyms "penetrating" and "irritating" both enter the Top 1.5\%, indicating that the latent space successfully bridges symbolic differences to align with authentic perceptual experiences. The model tends to identify more discriminative and specific descriptors, such as "spicy" (1.842\%) for Asarone and "gasoline" (1.473\%) for 1,3-Butadiene ranking prominently, while "aromatic" ranks moderately. This "specificity-oriented" tendency offers greater retrieval value in practical applications.
\begin{table*}
  \centering
  \small
  \begin{tabular}{llp{3.5cm}p{5cm}}
    \toprule
    \textbf{Name} & \textbf{CAS} & \textbf{Description (PubChem)} & \textbf{Retrieved Descriptors (Rank, Top\%)} \\
    \midrule
    \multicolumn{4}{l}{\textit{Zero-shot}} \\
    \midrule
    Bromobenzene & 108-86-1 & AROMATIC ODOR; Pungent odor & \textbf{penetrating (14/1086, 1.3\%)} \newline \textbf{irritating (17/1086, 1.6\%)} \newline pungent (111/1086, 10.2\%) \newline aromatic (258/1086, 23.8\%) \\
    \hdashline
    Iodoform & 75-47-8 & Characteristic, disagreeable odor; pungent, disagreeable odor & \textbf{irritating (11/1086, 1.0\%)} \newline \textbf{penetrating (13/1086, 1.2\%)} \newline \textbf{repulsive (46/1086, 4.2\%)} \newline pungent (117/1086, 10.8\%) \newline disagreeable (163/1086, 15.0\%) \\
    \hdashline
    Cyclohexylamine & 108-91-8 & Strong, fishy, amine odor & \textbf{amine (1/1086, 0.1\%)} \newline \textbf{ammoniacal (2/1086, 0.2\%)} \newline \textbf{ammonia (3/1086, 0.3\%)} \newline \textbf{dead animal (6/1086, 0.6\%)} \newline \textbf{rotten fish (11/1086, 1.0\%)} \newline \textbf{fishy (21/1086, 1.9\%)} \\
    \hdashline
    2,4-Dinitrotoluene & 121-14-2 & Slight odor & \textbf{odorless (1/1086, 0.1\%)} \newline \textbf{slight (2/1086, 0.2\%)} \newline \textbf{neutral (3/1086, 0.3\%)} \newline \textbf{weak (5/1086, 0.5\%)} \\
    \hdashline
    D-Glucose & - & Odorless & \textbf{odorless (1/1086, 0.1\%)} \newline \textbf{neutral (5/1086, 0.5\%)} \newline \textbf{slight (7/1086, 0.6\%)} \newline \textbf{weak (9/1086, 0.8\%)} \\
    \hdashline
    Estriol & 50-27-1 & Odorless & \textbf{odorless (1/1086, 0.1\%)} \newline \textbf{neutral (3/1086, 0.3\%)} \newline \textbf{slight (4/1086, 0.4\%)} \newline \textbf{weak (29/1086, 2.7\%)} \\
    \midrule
    \multicolumn{4}{l}{\textit{Strict Zero-shot}} \\
    \midrule
    Asarone & 2883-98-9 & Yellow to medium brown, moderately viscous liquid with a pleasant, spicy aromatic odor. & \textbf{spicy (20/1086, 1.8\%)} \newline aromatic (75/1086, 6.9\%) \\
    \hdashline
    1,3-Butadiene & 106-99-0 & Mild aromatic or gasoline-like odor & \textbf{gasoline (16/1086, 1.5\%)} \newline aromatic (434/1086, 40.0\%) \\
    \hdashline
    Propylene & - & Practically odorless; aromatic; Faint, petroleum-like & \textbf{gasoline (11/1086, 1.0\%)} \newline \textbf{light (39/1086, 3.6\%)} \newline petroleum (125/1086, 11.5\%) \newline aromatic (487/1086, 44.8\%) \\
    \hdashline
    Ethylene & - & Sweet; Olefinic, hedonic tone: unpleasant to neutral & \textbf{hedonic (21/1086, 1.9\%)} \newline unpleasant (133/1086, 12.2\%) \newline neutral (187/1086, 17.2\%) \newline sweet (427/1086, 39.3\%) \\
    \hdashline
    Formaldehyde & 50-00-0 & Pungent, suffocating odor; Pungent, irritating odor  & \textbf{irritating (1/1086, 0.1\%)} \newline \textbf{pungent (34/1086, 3.1\%)} \newline suffocating (55/1086, 5.1\%) \\
    \bottomrule
  \end{tabular}
  \caption{\label{appendix_odor_retrieval}
    Detailed zero-shot SMILES-to-odor retrieval results. Rank indicates the position among 1,086 candidate descriptors (lower is better). Results in the top 5\% are shown in \textbf{bold}. 
  }
\end{table*}

\subsubsection{smiles-to-receptor}
We select molecule-receptor pairs with explicitly reported "activating" or "non-activating" relationships from the literature as ground truth to evaluate the model's retrieval ranking capability for positive samples and its rejection capability for negative samples. The molecule-receptor pairs in the test set have never appeared in the training set. Results are presented in Table~\ref{appendix_receptor_retrieval}. The model demonstrates exceptionally high accuracy in retrieving the OR5A2 receptor and its ligands (primarily macrocyclic musk molecules, MCM), with all rankings at 2 (Rank 1 being the corresponding ligands present in the training set). Although the training set contains MCM molecules and OR5A2 sequences, they never appeared as pairs. The model successfully extracted the shared structural features of MCM and the sequence features of OR5A2, correctly mapping them in the latent space, demonstrating its capability to understand the correspondence between chemical structural families and receptor families. The model also exhibits good generalization across other chemical families, while revealing differences in modeling difficulty among receptor families. The propionate ion retrieval ranking is relatively lower (26.73\%), possibly due to the complexity of small molecular ion representation and data scarcity. All "non-activating" samples exhibit retrieval rankings significantly lower than "activating" samples, predominantly distributed in the 30\%-80\% range. This distributional difference indicates that the latent space constructed by the model not only brings positive sample pairs closer together but also effectively pushes negative sample pairs apart, demonstrating reliable value for biological screening. 
\begin{table*}
  \centering
  \small
  % Requires: \usepackage{arydshln}
  \begin{tabular}{lllp{3cm}c}
    \toprule
    \textbf{Name} & \textbf{CAS} & \textbf{Gene} & \textbf{Source} & \textbf{ (Rank, Top\%)} \\
    \midrule
    \multicolumn{5}{l}{\textit{Activated — Macrocyclic Musks (MCM)}} \\
    \midrule
    Globanone & 3100-36-5 & OR5A2 & \cite{yoshikawa2022odorant} & \textbf{(2/636, 0.315\%)} \\
    \hdashline
    Muscenone delta & 82356-51-2 & OR5A2 & \cite{yoshikawa2022odorant} & \textbf{(2/636, 0.315\%)} \\
    \hdashline
    Cosmone & 259854-70-1 & OR5A2 & \cite{yoshikawa2022odorant} & \textbf{(2/636, 0.315\%)} \\
    \hdashline
    Cyclopentadecanol & 4727-17-7 & OR5A2 & \cite{yoshikawa2022odorant} & \textbf{(2/636, 0.315\%)} \\
    \hdashline
    Ambrettolide & 7779-50-2 & OR5A2 & \cite{yoshikawa2022odorant} & \textbf{(2/636, 0.315\%)} \\
    \hdashline
    Habanolide & 111879-80-2 & OR5A2 & \cite{yoshikawa2022odorant} & \textbf{(2/636, 0.315\%)} \\
    \midrule
    \multicolumn{5}{l}{\textit{Activated — Other Molecules}} \\
    \midrule
    $\omega$-Dodecanolactam & 947-04-6 & OR5A2 & \cite{yoshikawa2022odorant} & \textbf{(2/636, 0.315\%)} \\
    \hdashline
    Trimethylamine & 75-50-3 & TAAR5 & \cite{gisladottir2020sequence,wallrabenstein2013human} & \textbf{(3/636, 0.472\%)} \\
    \hdashline
    Romandolide & 236391-76-7 & OR5A2 & \cite{yoshikawa2022odorant} & \textbf{(10/636, 1.572\%)} \\
    \hdashline
    Amber xtreme & 476332-65-7 & OR5A2 & \cite{yoshikawa2022odorant} & \textbf{(11/636, 1.730\%)} \\
    \hdashline
    butane-2,3-dione & 431-03-8 & OR52H1 & \cite{geithe2015butter,zeng2023molecular} & \textbf{(15/636, 2.359\%)} \\
    \hdashline
    butane-2,3-dione & 431-03-8 & OR51B5 & \cite{geithe2015butter,zeng2023molecular} & (65/636, 10.220\%) \\
    \hdashline
    Propionate ion & 72-03-7 & OR51E2 & \cite{billesbolle2023structural,choi2023understanding} & (170/636, 26.730\%) \\
    \midrule
    \multicolumn{5}{l}{\textit{Inactivated}} \\
    \midrule
    (E)-2-Decenal & 3913-81-3 & OR2W1 & \cite{haag2022key} & (246/636, 38.679\%) \\
    \hdashline
    $\delta$-Dodecalactone & 713-95-1 & OR5A2 & \cite{yoshikawa2022odorant} & (366/636, 57.547\%) \\
    \hdashline
    2-Propionyl-1-pyrroline & 133447-37-7 & OR2W1 & \cite{haag2022key} & (394/636, 61.950\%) \\
    \hdashline
    Raspberry ketone & 5471-51-2 & OR5A2 & \cite{yoshikawa2022odorant} & (478/636, 75.157\%) \\
    \hdashline
    p-Cresyl phenyl acetate & 101-94-0 & OR5A2 & \cite{yoshikawa2022odorant} & (499/636, 78.459\%) \\
    \bottomrule
  \end{tabular}
  \caption{
    Detailed zero-shot SMILES-to-receptor retrieval results. Rank\% indicates the percentile ranking among 636 candidate receptors. For activated pairs, lower is better (results below 5\% in \textbf{bold}). For inactivated pairs, higher rank\% indicates successful rejection.
  }
  \label{appendix_receptor_retrieval}
\end{table*}

\begin{table*}[t]
  \centering
  \begin{tabular}{lcccc}
    \toprule
    & \multicolumn{4}{c}{\textbf{Thresholds (Abraham)}} \\
    \cmidrule(lr){2-5}
    \textbf{Method} & R$^2$ $\uparrow$ & MAE $\downarrow$ & Pearson $\uparrow$ & MSE $\downarrow$ \\
    \midrule
    GCN & -0.304(0.431) & 1.309(0.247) & 0.220(0.089) & 2.912(0.963) \\
    AttentiveFP & 0.102(0.634) & 1.073(0.375) & 0.591(0.351) & 2.006(1.417) \\
    GIN & 0.339(0.135) & 0.969(0.129) & 0.723(0.018) & 1.477(0.300) \\
    Morgan & 0.360(0.215) & 0.924(0.119) & 0.596(0.175) & 1.431(0.481) \\
    POM & 0.565(0.087) & 0.731(0.082) & 0.788(0.024) & 0.972(0.195) \\
    Uni-Mol & 0.581(0.064) & 0.770(0.091) & 0.779(0.047) & 0.936(0.142) \\
    ChemBERTa & 0.577(0.051) & 0.757(0.071) & 0.814(0.022) & 0.944(0.114) \\
    NOSE & \textbf{0.652(0.072)} & \textbf{0.711(0.083)} & \textbf{0.836(0.026)} & \textbf{0.778(0.161)} \\
    \bottomrule
  \end{tabular}
  \caption{Basic perceptual attribute prediction: Thresholds.}
  \label{Complete_metrics_1}
\end{table*}

\begin{table*}[t]
  \centering
  \begin{tabular}{lcccc}
    \toprule
    & \multicolumn{4}{c}{\textbf{Pleasantness (Keller)}} \\
    \cmidrule(lr){2-5}
    \textbf{Method} & R$^2$ $\uparrow$ & MAE $\downarrow$ & Pearson $\uparrow$ & MSE $\downarrow$ \\
    \midrule
    GIN & -0.175(0.081) & 10.540(0.178) & 0.299(0.037) & 205.154(14.095) \\
    Uni-Mol & -0.071(0.204) & 10.576(1.354) & 0.680(0.017) & 186.939(35.551) \\
    GCN & 0.073(0.025) & 8.610(0.128) & 0.435(0.007) & 161.831(4.433) \\
    Morgan & 0.180(0.116) & 9.280(0.879) & 0.515(0.034) & 143.160(20.192) \\
    ChemBERTa & 0.120(0.216) & 9.194(0.675) & 0.649(0.036) & 153.589(37.664) \\
    AttentiveFP & 0.231(0.075) & 7.712(0.106) & 0.518(0.051) & 134.211(13.115) \\
    POM & 0.405(0.022) & 7.139(0.220) & 0.681(0.026) & 103.903(3.857) \\
    NOSE & \textbf{0.488(0.074)} & \textbf{6.911(0.609)} & \textbf{0.715(0.050)} & \textbf{89.280(12.934)} \\
    \bottomrule
  \end{tabular}
  \caption{Basic perceptual attribute prediction: Pleasantness (Keller).}
  \label{Complete_metrics_2}
\end{table*}

\begin{table*}[t]
  \centering
  \begin{tabular}{lcccc}
    \toprule
    & \multicolumn{4}{c}{\textbf{Pleasantness (Sagar)}} \\
    \cmidrule(lr){2-5}
    \textbf{Method} & R$^2$ $\uparrow$ & MAE $\downarrow$ & Pearson $\uparrow$ & MSE $\downarrow$ \\
    \midrule
    GIN & -0.139(0.092) & 0.497(0.021) & 0.079(0.056) & 0.341(0.028) \\
    Morgan & -0.085(0.054) & 0.489(0.010) & 0.213(0.055) & 0.325(0.016) \\
    GCN & -0.053(0.100) & 0.465(0.024) & 0.116(0.272) & 0.316(0.030) \\
    ChemBERTa & -0.028(0.070) & 0.475(0.012) & 0.152(0.125) & 0.308(0.021) \\
    AttentiveFP & -0.008(0.030) & 0.465(0.007) & 0.233(0.046) & 0.302(0.009) \\
    POM & 0.004(0.050) & 0.471(0.017) & 0.262(0.020) & 0.299(0.015) \\
    Uni-Mol & 0.030(0.170) & 0.456(0.039) & 0.144(0.335) & 0.291(0.051) \\
    NOSE & \textbf{0.105(0.064)} & \textbf{0.447(0.011)} & \textbf{0.397(0.020)} & \textbf{0.268(0.019)} \\
    \bottomrule
  \end{tabular}
  \caption{Basic perceptual attribute prediction: Pleasantness (Sagar).}
  \label{Complete_metrics_3}
\end{table*}

\begin{table*}[t]
  \centering
  \begin{tabular}{lcccc}
    \toprule
    & \multicolumn{4}{c}{\textbf{Intensity (Keller)}} \\
    \cmidrule(lr){2-5}
    \textbf{Method} & R$^2$ $\uparrow$ & MAE $\downarrow$ & Pearson $\uparrow$ & MSE $\downarrow$ \\
    \midrule
    Uni-Mol & -0.470(0.317) & 17.189(1.927) & 0.270(0.031) & 470.285(101.497) \\
    GIN & -0.318(0.165) & 16.295(0.820) & 0.063(0.141) & 421.540(52.643) \\
    GCN & -0.249(0.107) & 16.457(0.641) & 0.109(0.052) & 399.513(34.145) \\
    Morgan& -0.248(0.053) & 15.981(0.431) & 0.053(0.024) & 399.332(16.902) \\
    ChemBERTa & -0.117(0.145) & 15.001(1.156) & 0.393(0.070) & 357.445(46.307) \\
    POM & -0.053(0.094) & 14.769(0.353) & 0.317(0.069) & 336.787(30.179) \\
    AttentiveFP & 0.054(0.025) & 14.234(0.156) & 0.326(0.012) & 302.630(8.013) \\
    NOSE & \textbf{0.156(0.020)} & \textbf{12.932(0.230)} & \textbf{0.418(0.010)} & \textbf{269.902(6.526)} \\
    \bottomrule
  \end{tabular}
  \caption{Basic perceptual attribute prediction: Intensity (Keller).}
  \label{Complete_metrics_4}
\end{table*}

\begin{table*}[t]
  \centering
  \begin{tabular}{lcccc}
    \toprule
    & \multicolumn{4}{c}{\textbf{Intensity (Sagar)}} \\
    \cmidrule(lr){2-5}
    \textbf{Method} & R$^2$ $\uparrow$ & MAE $\downarrow$ & Pearson $\uparrow$ & MSE $\downarrow$ \\
    \midrule
    GIN & -1.492(1.869) & 0.540(0.223) & -0.056(0.266) & 0.508(0.381) \\
    Morgan& -0.584(0.191) & 0.465(0.026) & 0.230(0.102) & 0.323(0.039) \\
    ChemBERTa & -0.336(0.136) & 0.462(0.022) & 0.447(0.082) & 0.272(0.028) \\
    AttentiveFP & -0.216(0.201) & 0.450(0.048) & 0.228(0.055) & 0.248(0.041) \\
    Uni-Mol & -0.317(0.102) & 0.442(0.022) & 0.372(0.051) & 0.269(0.021) \\
    POM & -0.076(0.134) & 0.383(0.031) & 0.334(0.061) & 0.220(0.027) \\
    GCN & 0.065(0.015) & \textbf{0.351(0.006)} & 0.305(0.042) & 0.191(0.003) \\
    NOSE & \textbf{0.120(0.109)} & 0.355(0.041) & \textbf{0.468(0.075)} & \textbf{0.179(0.022)} \\
    \bottomrule
  \end{tabular}
  \caption{Basic perceptual attribute prediction: Intensity (Sagar).}
  \label{Complete_metrics_5}
\end{table*}

\begin{table*}[t]
  \centering
  \begin{tabular}{lcccc}
    \toprule
    & \multicolumn{4}{c}{\textbf{Intensity (Ravia)}} \\
    \cmidrule(lr){2-5}
    \textbf{Method} & R$^2$ $\uparrow$ & MAE $\downarrow$ & Pearson $\uparrow$ & MSE $\downarrow$ \\
    \midrule
    Morgan& -2.832(4.010) & 24.444(17.045) & 0.103(0.186) & 1062.533(1111.892) \\
    Uni-Mol & -0.407(0.352) & 15.176(1.993) & 0.307(0.024) & 390.301(97.745) \\
    GIN & -0.191(0.172) & 14.514(1.228) & 0.188(0.114) & 330.250(47.774) \\
    GCN & -0.109(0.006) & 13.426(0.540) & 0.192(0.014) & 307.666(1.719) \\
    POM & -0.080(0.057) & 12.833(0.297) & 0.319(0.055) & 299.530(15.798) \\
    ChemBERTa & 0.062(0.147) & 12.068(0.881) & 0.471(0.064) & 260.144(40.728) \\
    AttentiveFP & 0.148(0.008) & 11.527(0.108) & 0.471(0.004) & 236.190(2.315) \\
    NOSE & \textbf{0.220(0.025)} & \textbf{11.078(0.351)} & \textbf{0.485(0.031)} & \textbf{216.326(6.824)} \\
    \bottomrule
  \end{tabular}
  \caption{Basic perceptual attribute prediction: Intensity (Ravia).}
  \label{Complete_metrics_6}
\end{table*}

\begin{table*}[t]
  \centering
  \begin{tabular}{l|ccc}
    \hline
    & \multicolumn{3}{c}{\textbf{GS-LF Multi-label Multi-class Classification}} \\
    \cmidrule(lr){2-4}
    \textbf{Method} & AUC $\uparrow$ & AUPRC $\uparrow$ & MCC $\uparrow$ \\
    \hline
    GIN & 0.856(0.001) & 0.270(0.008) & 0.064(0.007) \\
    GCN & 0.858(0.001) & 0.279(0.001) & 0.084(0.003) \\
    Morgan& 0.850(0.002) & 0.320(0.002) & 0.222(0.023) \\
    AttentiveFP & 0.867(0.004) & 0.328(0.012) & 0.203(0.026) \\
    POM & 0.868(0.002) & 0.336(0.005) & 0.233(0.023) \\
    ChemBERTa & 0.875(0.001) & 0.342(0.005) & 0.240(0.016) \\
    Uni-Mol & 0.873(0.001) & 0.347(0.005) & 0.262(0.020) \\
    NOSE & \textbf{0.876(0.001)} & \textbf{0.351(0.002)} & \textbf{0.268(0.010)} \\
    \hline
  \end{tabular}
  \caption{Semantic description prediction on GS-LF dataset.}
  \label{Complete_metrics_7}
\end{table*}

\begin{table*}[t]
  \centering
  \begin{tabular}{lcccc}
    \toprule
    & \multicolumn{4}{c}{\textbf{Multi-label Regression (Keller)}} \\
    \cmidrule(lr){2-5}
    \textbf{Method} & R$^2$ $\uparrow$ & MAE $\downarrow$ & Pearson $\uparrow$ & MSE $\downarrow$ \\
    \midrule
    GIN & -0.229(0.072) & 6.528(0.139) & 0.128(0.007) & 90.903(4.613) \\
    Uni-Mol & -0.178(0.060) & 6.741(0.209) & 0.330(0.050) & 90.373(5.967) \\
    Morgan& -0.087(0.024) & 6.483(0.015) & 0.244(0.023) & 80.531(1.690) \\
    GCN & -0.106(0.011) & 6.247(0.031) & 0.171(0.006) & 80.182(0.243) \\
    POM & -0.125(0.076) & 6.304(0.258) & 0.314(0.039) & 78.649(4.872) \\
    AttentiveFP & -0.143(0.171) & 6.232(0.647) & 0.327(0.065) & 80.149(14.116) \\
    ChemBERTa & 0.018(0.056) & 6.110(0.255) & 0.330(0.089) & 71.841(5.117) \\
    NOSE & \textbf{0.075(0.040)} & \textbf{5.862(0.225)} & \textbf{0.348(0.060)} & \textbf{67.161(4.161)} \\
    \bottomrule
  \end{tabular}
  \caption{Semantic description prediction on Keller dataset.}
  \label{Complete_metrics_8}
\end{table*}

\begin{table*}[t]
  \centering
  \begin{tabular}{lcccc}
    \toprule
    & \multicolumn{4}{c}{\textbf{Multi-label Regression (Sagar)}} \\
    \cmidrule(lr){2-5}
    \textbf{Method} & R$^2$ $\uparrow$ & MAE $\downarrow$ & Pearson $\uparrow$ & MSE $\downarrow$ \\
    \midrule
    POM & -0.835(0.821) & 0.405(0.113) & 0.065(0.062) & 0.281(0.094) \\
    GCN & -0.933(1.067) & 0.407(0.098) & 0.108(0.085) & 0.254(0.076) \\
    GIN & -0.654(0.022) & 0.358(0.005) & -0.034(0.016) & 0.248(0.006) \\
    Morgan& -0.563(0.273) & 0.395(0.018) & 0.008(0.076) & 0.233(0.019) \\
    Uni-Mol & -0.677(0.218) & 0.355(0.009) & 0.116(0.042) & 0.252(0.002) \\
    AttentiveFP & -0.481(0.053) & 0.359(0.012) & 0.011(0.013) & 0.230(0.009) \\
    ChemBERTa & \textbf{-0.275(0.057)} & 0.376(0.018) & 0.105(0.051) & \textbf{0.218(0.006)} \\
    NOSE & -0.305(0.106) & \textbf{0.343(0.017)} & \textbf{0.123(0.068)} & 0.225(0.017) \\
    \bottomrule
  \end{tabular}
  \caption{Semantic description prediction on Sagar dataset.}
  \label{Complete_metrics_9}
\end{table*}

\begin{table*}[t]
  \centering
  \begin{tabular}{lcccc}
    \toprule
    & \multicolumn{4}{c}{\textbf{Mixture Intensity}} \\
    \cmidrule(lr){2-5}
    \textbf{Method} & R$^2$ $\uparrow$ & MAE $\downarrow$ & Pearson $\uparrow$ & MSE $\downarrow$ \\
    \midrule
    GIN & -2.821(0.297) & 0.857(0.017) & 0.006(0.143) & 1.056(0.082) \\
    GCN & -0.455(0.503) & 0.516(0.094) & 0.491(0.073) & 0.402(0.139) \\
    Morgan& 0.143(0.307) & 0.397(0.081) & 0.575(0.063) & 0.237(0.085) \\
    AttentiveFP & 0.260(0.059) & 0.361(0.006) & 0.597(0.034) & 0.205(0.016) \\
    Uni-Mol & 0.319(0.106) & 0.355(0.028) & 0.637(0.069) & 0.188(0.029) \\
    POM & 0.347(0.061) & 0.359(0.018) & 0.603(0.033) & 0.180(0.017) \\
    ChemBERTa & 0.326(0.023) & 0.350(0.005) & 0.618(0.005) & 0.186(0.006) \\
    NOSE & \textbf{0.389(0.052)} & \textbf{0.333(0.021)} & \textbf{0.657(0.029)} & \textbf{0.169(0.014)} \\
    \bottomrule
  \end{tabular}
  \caption{Mixture prediction tasks: Intensity.}
  \label{Complete_metrics_10}
\end{table*}

\begin{table*}[t]
  \centering
  \begin{tabular}{lcccc}
    \toprule
    & \multicolumn{4}{c}{\textbf{Mixture Pleasantness}} \\
    \cmidrule(lr){2-5}
    \textbf{Method} & R$^2$ $\uparrow$ & MAE $\downarrow$ & Pearson $\uparrow$ & MSE $\downarrow$ \\
    \midrule
    GCN & -14.296(9.524) & 3.685(1.723) & 0.332(0.418) & 17.477(10.883) \\
    GIN & -0.096(0.123) & 0.879(0.056) & 0.473(0.029) & 1.252(0.140) \\
    ChemBERTa & 0.452(0.074) & 0.641(0.060) & 0.793(0.009) & 0.626(0.084) \\
    AttentiveFP & 0.490(0.068) & 0.616(0.051) & 0.744(0.015) & 0.583(0.078) \\
    Morgan& 0.498(0.187) & 0.599(0.113) & 0.814(0.012) & 0.574(0.214) \\
    Uni-Mol & 0.509(0.047) & 0.614(0.023) & 0.795(0.035) & 0.561(0.054) \\
    POM & 0.557(0.029) & 0.561(0.002) & 0.771(0.018) & 0.506(0.033) \\
    NOSE & \textbf{0.636(0.047)} & \textbf{0.534(0.033)} & \textbf{0.846(0.012)} & \textbf{0.416(0.054)} \\
    \bottomrule
  \end{tabular}
  \caption{Mixture prediction tasks: Pleasantness.}
  \label{Complete_metrics_11}
\end{table*}

\begin{table*}[t]
  \centering
  \small
  \begin{tabular}{lcccc}
    \toprule
    & \multicolumn{4}{c}{\textbf{Thresholds (Abraham)}} \\
    \cmidrule(lr){2-5}
    \textbf{Method} & R$^2$ $\uparrow$ & MAE $\downarrow$ & Pearson $\uparrow$ & MSE $\downarrow$ \\
    \midrule
    Adapter (Desc: 10.00M, Rec: 10.00M) & 0.161(0.616) & 1.074(0.421) & 0.768(0.049) & 1.875(1.375) \\
    Adapter (Desc: 29.17M, Rec: 10.00M) & 0.331(0.179) & 0.986(0.148) & 0.742(0.042) & 1.494(0.401) \\
    Inter & 0.338(0.446) & 0.978(0.374) & 0.782(0.063) & 1.478(0.995) \\
    Hard + Soft $\lambda$=1.0 & 0.440(0.072) & 0.884(0.085) & 0.749(0.006) & 1.252(0.162) \\
    Receptor only & 0.444(0.173) & 0.909(0.173) & 0.791(0.041) & 1.242(0.387) \\
    Hard + Soft $\lambda$=0.5 & 0.480(0.037) & 0.887(0.034) & 0.771(0.020) & 1.162(0.083) \\
    Only Hard & 0.449(0.261) & 0.891(0.246) & 0.810(0.030) & 1.232(0.582) \\
    Hard + Soft $\lambda$=0.1 & 0.538(0.097) & 0.827(0.088) & 0.764(0.056) & 1.032(0.216) \\
    Only Soft $\lambda$=2.0 & 0.515(0.050) & 0.868(0.031) & 0.795(0.008) & 1.083(0.111) \\
    No Orthogonal & 0.506(0.254) & 0.849(0.226) & 0.804(0.022) & 1.104(0.567) \\
    Molecule only & 0.581(0.064) & 0.770(0.091) & 0.779(0.047) & 0.936(0.142) \\
    Inter + Weak & 0.570(0.100) & 0.792(0.131) & 0.813(0.004) & 0.961(0.223) \\
    Inter + Intra & 0.582(0.140) & 0.779(0.158) & 0.802(0.031) & 0.934(0.313) \\
    Description only & 0.608(0.026) & 0.746(0.038) & 0.816(0.025) & 0.875(0.058) \\
    NOSE & \textbf{0.652(0.072)} & \textbf{0.711(0.083)} & \textbf{0.836(0.026)} & \textbf{0.778(0.161)} \\
    \bottomrule
  \end{tabular}
  \caption{Ablation study: Basic perceptual attribute prediction: Thresholds.}
  \label{tab:ablation_1}
\end{table*}

\begin{table*}[t]
  \centering
  \small
  \begin{tabular}{lcccc}
    \toprule
    & \multicolumn{4}{c}{\textbf{Pleasantness (Keller)}} \\
    \cmidrule(lr){2-5}
    \textbf{Method} & R$^2$ $\uparrow$ & MAE $\downarrow$ & Pearson $\uparrow$ & MSE $\downarrow$ \\
    \midrule
    No Orthogonal & 0.343(0.078) & 7.818(0.487) & 0.622(0.046) & 114.722(13.697) \\
    Hard + Soft $\lambda$=1.0 & 0.354(0.055) & 8.079(0.355) & 0.623(0.040) & 112.703(9.565) \\
    Molecule only & -0.071(0.204) & 10.576(1.354) & 0.680(0.017) & 186.939(35.551) \\
    Description only & 0.366(0.110) & 7.959(0.877) & 0.647(0.063) & 110.652(19.235) \\
    Receptor only & 0.368(0.104) & 8.025(0.690) & 0.636(0.102) & 110.308(18.106) \\
    Inter + Weak & 0.383(0.090) & 7.751(0.422) & 0.654(0.063) & 107.715(15.682) \\
    Inter + Intra & 0.401(0.066) & 7.632(0.518) & 0.648(0.039) & 104.572(11.437) \\
    Adapter (Desc: 10.00M, Rec: 10.00M) & 0.378(0.117) & 7.393(0.688) & 0.669(0.030) & 108.561(20.457) \\
    Hard + Soft $\lambda$=0.5 & 0.398(0.018) & 7.681(0.457) & 0.663(0.025) & 105.027(3.067) \\
    Adapter (Desc: 29.17M, Rec: 10.00M) & 0.420(0.039) & 7.486(0.332) & 0.652(0.032) & 101.285(6.863) \\
    Hard + Soft $\lambda$=0.1 & 0.432(0.075) & 7.197(0.638) & 0.662(0.062) & 99.119(13.024) \\
    Only Soft $\lambda$=2.0 & 0.441(0.104) & 7.087(0.816) & 0.671(0.084) & 97.555(18.119) \\
    Only Hard & 0.468(0.097) & 7.004(0.368) & 0.704(0.077) & 92.803(16.947) \\
    NOSE & 0.488(0.074) & 6.911(0.609) & 0.715(0.050) & 89.280(12.934) \\
    Inter & \textbf{0.520(0.155)} & \textbf{6.878(0.967)} & \textbf{0.734(0.096)} & \textbf{83.762(27.123)} \\
    \bottomrule
  \end{tabular}
  \caption{Ablation study: Pleasantness prediction (Keller).}
  \label{tab:ablation_2}
\end{table*}

\begin{table*}[t]
  \centering
  \small
  \begin{tabular}{lcccc}
    \toprule
    & \multicolumn{4}{c}{\textbf{Pleasantness (Sagar)}} \\
    \cmidrule(lr){2-5}
    \textbf{Method} & R$^2$ $\uparrow$ & MAE $\downarrow$ & Pearson $\uparrow$ & MSE $\downarrow$ \\
    \midrule
    Inter + Intra & -0.282(0.336) & 0.511(0.047) & 0.025(0.245) & 0.384(0.101) \\
    Inter & -0.219(0.214) & 0.505(0.023) & -0.034(0.285) & 0.365(0.064) \\
    Adapter (Desc: 10.00M, Rec: 10.00M) & -0.208(0.152) & 0.492(0.036) & -0.127(0.181) & 0.362(0.046) \\
    Only Hard & -0.169(0.027) & 0.508(0.016) & 0.085(0.158) & 0.350(0.008) \\
    No Orthogonal & -0.148(0.110) & 0.490(0.022) & 0.083(0.078) & 0.344(0.033) \\
    Hard + Soft $\lambda$=0.5 & -0.110(0.163) & 0.484(0.029) & 0.011(0.166) & 0.333(0.049) \\
    Adapter (Desc: 29.17M, Rec: 10.00M) & -0.083(0.167) & 0.477(0.031) & 0.129(0.329) & 0.325(0.050) \\
    Hard + Soft $\lambda$=1.0 & -0.108(0.374) & 0.474(0.061) & 0.298(0.156) & 0.332(0.112) \\
    Receptor only & -0.103(0.031) & 0.467(0.021) & 0.188(0.056) & 0.331(0.009) \\
    Hard + Soft $\lambda$=0.1 & -0.104(0.379) & 0.476(0.069) & 0.362(0.099) & 0.331(0.114) \\
    Inter + Weak & 0.001(0.100) & 0.476(0.022) & 0.229(0.154) & 0.299(0.030) \\
    Description only & -0.009(0.105) & 0.463(0.011) & 0.209(0.122) & 0.302(0.032) \\
    Molecule only & 0.030(0.170) & 0.456(0.039) & 0.144(0.335) & 0.291(0.051) \\
    Only Soft $\lambda$=2.0 & 0.064(0.099) & 0.451(0.021) & 0.223(0.249) & 0.281(0.030) \\
    NOSE & \textbf{0.105(0.064)} & \textbf{0.447(0.011)} & \textbf{0.397(0.020)} & \textbf{0.268(0.019)} \\
    \bottomrule
  \end{tabular}
  \caption{Ablation study: Basic perceptual attribute prediction: Pleasantness (Sagar).}
  \label{tab:ablation_3}
\end{table*}

\begin{table*}[t]
  \centering
  \small
  \begin{tabular}{lcccc}
    \toprule
    & \multicolumn{4}{c}{\textbf{Intensity (Keller)}} \\
    \cmidrule(lr){2-5}
    \textbf{Method} & R$^2$ $\uparrow$ & MAE $\downarrow$ & Pearson $\uparrow$ & MSE $\downarrow$ \\
    \midrule
    Molecule only & -0.470(0.317) & 17.189(1.927) & 0.270(0.031) & 470.285(101.497) \\
    Inter & -0.224(0.096) & 16.201(0.767) & 0.206(0.042) & 391.690(30.734) \\
    Hard + Soft $\lambda$=0.5 & -0.097(0.141) & 15.088(0.689) & 0.246(0.074) & 351.066(44.974) \\
    Adapter (Desc: 29.17M, Rec: 10.00M) & -0.103(0.194) & 14.797(1.517) & 0.304(0.090) & 352.751(62.021) \\
    Hard + Soft $\lambda$=1.0 & -0.024(0.165) & 14.839(1.500) & 0.322(0.117) & 327.603(52.872) \\
    Hard + Soft $\lambda$=0.1 & 0.020(0.052) & 14.348(0.501) & 0.320(0.036) & 313.497(16.783) \\
    Only Hard & -0.015(0.068) & 14.513(0.815) & 0.336(0.025) & 324.564(21.719) \\
    Adapter (Desc: 10.00M, Rec: 10.00M) & 0.023(0.085) & 14.074(0.633) & 0.315(0.055) & 312.679(27.280) \\
    Description only & 0.026(0.013) & 14.231(0.442) & 0.321(0.029) & 311.741(4.179) \\
    Receptor only & 0.033(0.133) & 13.952(1.122) & 0.323(0.119) & 309.369(42.499) \\
    Inter + Weak & 0.043(0.091) & 14.029(0.819) & 0.329(0.075) & 306.240(29.163) \\
    No Orthogonal & 0.050(0.100) & 14.234(0.574) & 0.355(0.077) & 303.921(32.127) \\
    Only Soft $\lambda$=2.0 & 0.052(0.066) & 13.684(0.673) & 0.347(0.054) & 303.291(21.067) \\
    Inter + Intra & 0.060(0.032) & 14.207(0.238) & 0.381(0.042) & 300.820(10.274) \\
    NOSE & \textbf{0.156(0.020)} & \textbf{12.932(0.230)} & \textbf{0.418(0.010)} & \textbf{269.902(6.526)} \\
    \bottomrule
  \end{tabular}
  \caption{Ablation study: Basic perceptual attribute prediction: Intensity (Keller).}
  \label{tab:ablation_4}
\end{table*}

\begin{table*}[t]
  \centering
  \small
  \begin{tabular}{lcccc}
    \toprule
    & \multicolumn{4}{c}{\textbf{Intensity (Sagar)}} \\
    \cmidrule(lr){2-5}
    \textbf{Method} & R$^2$ $\uparrow$ & MAE $\downarrow$ & Pearson $\uparrow$ & MSE $\downarrow$ \\
    \midrule
    Inter & -0.389(0.244) & 0.455(0.034) & 0.360(0.132) & 0.283(0.050) \\
    Only Soft $\lambda$=2.0 & -0.253(0.355) & 0.455(0.054) & 0.347(0.180) & 0.256(0.072) \\
    Molecule only & -0.317(0.102) & 0.442(0.022) & 0.372(0.051) & 0.269(0.021) \\
    Hard + Soft $\lambda$=0.1 & -0.112(0.065) & 0.404(0.040) & 0.273(0.166) & 0.227(0.013) \\
    No Orthogonal & -0.138(0.253) & 0.398(0.072) & 0.308(0.108) & 0.232(0.052) \\
    Adapter (Desc: 29.17M, Rec: 10.00M) & -0.177(0.132) & 0.423(0.020) & 0.430(0.087) & 0.240(0.027) \\
    Hard + Soft $\lambda$=0.5 & -0.104(0.222) & 0.423(0.045) & 0.429(0.113) & 0.225(0.045) \\
    Description only & -0.072(0.142) & 0.390(0.059) & 0.378(0.173) & 0.219(0.029) \\
    Receptor only & -0.027(0.045) & 0.384(0.017) & 0.361(0.152) & 0.210(0.009) \\
    Hard + Soft $\lambda$=1.0 & -0.027(0.184) & 0.375(0.053) & 0.377(0.076) & 0.209(0.038) \\
    Only Hard & -0.021(0.162) & 0.391(0.020) & 0.417(0.068) & 0.208(0.033) \\
    Inter + Weak & 0.081(0.003) & \textbf{0.347(0.021)} & 0.327(0.039) & 0.188(0.001) \\
    Adapter (Desc: 10.00M, Rec: 10.00M) & 0.020(0.143) & 0.381(0.049) & 0.435(0.089) & 0.200(0.029) \\
    NOSE & 0.120(0.109) & 0.355(0.041) & 0.468(0.075) & 0.179(0.022) \\
    Inter + Intra & \textbf{0.177(0.202)} & 0.350(0.050) & \textbf{0.541(0.073)} & \textbf{0.168(0.041)} \\
    \bottomrule
  \end{tabular}
  \caption{Ablation study: Basic perceptual attribute prediction: Intensity (Sagar).}
  \label{tab:ablation_5}
\end{table*}

\begin{table*}[t]
  \centering
  \small
  \begin{tabular}{lcccc}
    \toprule
    & \multicolumn{4}{c}{\textbf{Intensity (Ravia)}} \\
    \cmidrule(lr){2-5}
    \textbf{Method} & R$^2$ $\uparrow$ & MAE $\downarrow$ & Pearson $\uparrow$ & MSE $\downarrow$ \\
    \midrule
    Molecule only & -0.407(0.352) & 15.176(1.993) & 0.307(0.024) & 390.301(97.745) \\
    Adapter (Desc: 29.17M, Rec: 10.00M) & 0.069(0.123) & 11.873(0.576) & 0.202(0.288) & 258.179(34.039) \\
    Receptor only & 0.058(0.135) & 11.988(0.719) & 0.347(0.134) & 261.213(37.444) \\
    Inter + Intra & 0.044(0.142) & 12.107(0.601) & 0.462(0.007) & 265.066(39.293) \\
    Adapter (Desc: 10.00M, Rec: 10.00M) & 0.079(0.074) & 12.012(0.130) & 0.426(0.119) & 255.522(20.592) \\
    Hard + Soft $\lambda$=0.1 & 0.082(0.131) & 11.788(0.637) & 0.421(0.093) & 254.596(36.293) \\
    Only Soft $\lambda$=2.0 & 0.113(0.121) & 11.761(0.608) & 0.312(0.238) & 245.860(33.608) \\
    Hard + Soft $\lambda$=0.5 & 0.090(0.112) & 11.746(0.577) & 0.477(0.008) & 252.269(30.948) \\
    Inter & 0.137(0.047) & 11.712(0.133) & 0.428(0.024) & 239.377(13.032) \\
    Hard + Soft $\lambda$=1.0 & 0.118(0.136) & 11.667(0.772) & 0.441(0.044) & 244.606(37.752) \\
    No Orthogonal & 0.155(0.107) & 11.550(0.548) & 0.468(0.036) & 234.259(29.573) \\
    Only Hard & 0.168(0.046) & 11.564(0.407) & 0.476(0.047) & 230.751(12.671) \\
    Description only & 0.207(0.026) & \textbf{11.056(0.206)} & 0.476(0.018) & 219.985(7.194) \\
    Inter + Weak & 0.210(0.067) & 11.205(0.815) & 0.480(0.055) & 219.046(18.540) \\
    NOSE & \textbf{0.220(0.025)} & 11.078(0.351) & \textbf{0.485(0.031)} & \textbf{216.326(6.824)} \\
    \bottomrule
  \end{tabular}
  \caption{Ablation study: Basic perceptual attribute prediction: Intensity (Ravia).}
  \label{tab:ablation_6}
\end{table*}

\begin{table*}[t]
  \centering
  \small
  \begin{tabular}{lccc}
    \toprule
    & \multicolumn{3}{c}{\textbf{GS-LF Multi-label Multi-class Classification}} \\
    \cmidrule(lr){2-4}
    \textbf{Method} & AUC $\uparrow$ & AUPRC $\uparrow$ & MCC $\uparrow$ \\
    \midrule
    No Orthogonal & 0.874(0.001) & 0.344(0.007) & 0.220(0.025) \\
    Adapter (Desc: 29.17M, Rec: 10.00M) & 0.875(0.003) & 0.344(0.005) & 0.250(0.024) \\
    Only Hard & 0.874(0.002) & 0.342(0.002) & 0.257(0.003) \\
    Only Soft $\lambda$=2.0 & 0.874(0.001) & 0.347(0.002) & 0.237(0.016) \\
    Inter + Weak & 0.875(0.001) & 0.345(0.004) & 0.241(0.027) \\
    Receptor only & 0.875(0.002) & 0.346(0.008) & 0.244(0.019) \\
    Molecule only & 0.873(0.001) & 0.347(0.005) & 0.262(0.020) \\
    Description only & 0.876(0.001) & 0.347(0.004) & 0.241(0.010) \\
    Hard + Soft $\lambda$=1.0 & 0.875(0.002) & 0.345(0.003) & 0.250(0.014) \\
    Hard + Soft $\lambda$=0.5 & 0.874(0.004) & 0.351(0.003) & 0.256(0.018) \\
    Hard + Soft $\lambda$=0.1 & 0.874(0.000) & 0.348(0.003) & 0.263(0.005) \\
    Adapter (Desc: 10.00M, Rec: 10.00M) & 0.874(0.004) & 0.348(0.009) & 0.261(0.010) \\
    Inter + Intra & 0.874(0.004) & 0.351(0.003) & 0.264(0.013) \\
    NOSE & 0.876(0.001) & 0.351(0.002) & \textbf{0.268(0.010)} \\
    Inter & \textbf{0.877(0.001)} & \textbf{0.354(0.004)} & 0.260(0.014) \\
    \bottomrule
  \end{tabular}
  \caption{Ablation study: semantic description prediction on GS-LF dataset.}
  \label{tab:ablation_7}
\end{table*}

\begin{table*}[t]
  \centering
  \small
  \begin{tabular}{lcccc}
    \toprule
    & \multicolumn{4}{c}{\textbf{Multi-label Regression (Keller)}} \\
    \cmidrule(lr){2-5}
    \textbf{Method} & $R^2$ $\uparrow$ & MAE $\downarrow$ & Pearson $r$ $\uparrow$ & MSE $\downarrow$ \\
    \midrule
    Receptor only & -0.093(0.060) & 6.369(0.226) & 0.233(0.061) & 79.899(3.491) \\
    Description only & -0.060(0.062) & 6.271(0.129) & 0.217(0.074) & 77.697(3.140) \\
    Hard + Soft $\lambda$=0.5 & -0.042(0.059) & 6.089(0.350) & 0.167(0.074) & 75.739(6.335) \\
    Molecule only & -0.178(0.060) & 6.741(0.209) & 0.330(0.050) & 90.373(5.967) \\
    Hard + Soft $\lambda$=1.0 & -0.034(0.038) & 6.136(0.141) & 0.243(0.097) & 75.666(4.221) \\
    Inter & -0.010(0.043) & 6.022(0.116) & 0.191(0.053) & 73.195(2.996) \\
    Only Hard & -0.038(0.069) & 5.993(0.084) & 0.246(0.047) & 75.320(4.707) \\
    Inter + Weak & 0.000(0.016) & 5.882(0.183) & 0.243(0.071) & 71.656(0.863) \\
    No Orthogonal & 0.001(0.016) & 6.069(0.244) & 0.265(0.106) & 71.798(3.617) \\
    Only Soft $\lambda$=2.0 & 0.006(0.023) & 5.977(0.147) & 0.248(0.064) & 71.690(1.685) \\
    Adapter (Desc: 29.17M, Rec: 10.00M) & 0.026(0.018) & 5.866(0.176) & 0.253(0.029) & 69.822(1.679) \\
    Inter + Intra & 0.016(0.033) & 5.760(0.083) & 0.258(0.011) & 70.365(2.269) \\
    Hard + Soft $\lambda$=0.1 & 0.028(0.031) & 5.815(0.269) & 0.254(0.052) & 69.549(3.590) \\
    Adapter (Desc: 10.00M, Rec: 10.00M) & 0.028(0.020) & \textbf{5.757(0.259)} & 0.277(0.081) & 69.049(2.518) \\
    NOSE & \textbf{0.075(0.040)} & 5.862(0.225) & \textbf{0.348(0.060)} & \textbf{67.161(4.161)} \\
    \bottomrule
  \end{tabular}
  \caption{Ablation study: semantic description prediction on Keller dataset.}
  \label{tab:ablation_8}
\end{table*}

\begin{table*}[t]
  \centering
  \small
  \begin{tabular}{lcccc}
    \toprule
    & \multicolumn{4}{c}{\textbf{Multi-label Regression (Sagar)}} \\
    \cmidrule(lr){2-5}
    \textbf{Method} & $R^2$ $\uparrow$ & MAE $\downarrow$ & Pearson $r$ $\uparrow$ & MSE $\downarrow$ \\
    \midrule
    Inter & -0.554(0.082) & 0.356(0.020) & 0.023(0.152) & 0.239(0.015) \\
    Inter + Weak & -0.389(0.012) & 0.343(0.001) & 0.007(0.108) & 0.233(0.006) \\
    Molecule only & -0.677(0.218) & 0.355(0.009) & 0.116(0.042) & 0.252(0.002) \\
    Hard + Soft $\lambda$=1.0 & -0.334(0.161) & 0.336(0.013) & -0.050(0.069) & 0.225(0.014) \\
    Inter + Intra & -0.364(0.158) & 0.339(0.017) & 0.042(0.176) & 0.226(0.015) \\
    Hard + Soft $\lambda$=0.5 & -0.388(0.176) & 0.339(0.014) & -0.024(0.085) & 0.224(0.015) \\
    Hard + Soft $\lambda$=0.1 & -0.423(0.228) & 0.332(0.022) & 0.007(0.148) & 0.224(0.018) \\
    No Orthogonal & -0.329(0.131) & 0.334(0.011) & -0.082(0.038) & 0.224(0.012) \\
    NOSE & -0.305(0.106) & 0.343(0.017) & 0.123(0.068) & 0.225(0.017) \\
    Receptor only & -0.338(0.152) & 0.329(0.011) & 0.075(0.056) & 0.227(0.009) \\
    Description only & -0.381(0.267) & 0.330(0.022) & 0.120(0.079) & 0.222(0.030) \\
    Adapter (Desc: 10.00M, Rec: 10.00M) & -0.298(0.113) & 0.329(0.006) & 0.075(0.152) & 0.224(0.007) \\
    Only Hard & -0.276(0.062) & 0.334(0.012) & \textbf{0.169(0.179)} & 0.224(0.016) \\
    Adapter (Desc: 29.17M, Rec: 10.00M) & -0.277(0.071) & 0.329(0.010) & 0.022(0.216) & \textbf{0.211(0.018)} \\
    Only Soft $\lambda$=2.0 & \textbf{-0.241(0.080)} & \textbf{0.327(0.005)} & 0.142(0.143) & 0.212(0.002) \\
    \bottomrule
  \end{tabular}
  \caption{Ablation study: semantic description prediction on Sagar dataset.}
  \label{tab:ablation_9}
\end{table*}

\begin{table*}[t]
  \centering
  \small
  \begin{tabular}{lcccc}
    \toprule
    & \multicolumn{4}{c}{\textbf{Mixture Intensity}} \\
    \cmidrule(lr){2-5}
    \textbf{Method} & $R^2$ $\uparrow$ & MAE $\downarrow$ & Pearson $r$ $\uparrow$ & MSE $\downarrow$ \\
    \midrule
    Receptor only & 0.224(0.085) & 0.379(0.010) & 0.582(0.012) & 0.215(0.023) \\
    Hard + Soft $\lambda$=0.1 & 0.258(0.095) & 0.360(0.028) & 0.530(0.093) & 0.205(0.026) \\
    Adapter (Desc: 10.00M, Rec: 10.00M) & 0.263(0.063) & 0.369(0.023) & 0.575(0.029) & 0.204(0.018) \\
    Inter + Weak & 0.285(0.049) & 0.362(0.006) & 0.566(0.061) & 0.198(0.013) \\
    Inter + Intra & 0.285(0.117) & 0.365(0.037) & 0.576(0.050) & 0.198(0.032) \\
    Description only & 0.285(0.086) & 0.361(0.017) & 0.549(0.070) & 0.198(0.024) \\
    Only Hard & 0.291(0.081) & 0.359(0.032) & 0.586(0.042) & 0.196(0.023) \\
    Adapter (Desc: 29.17M, Rec: 10.00M) & 0.335(0.098) & 0.356(0.029) & 0.611(0.041) & 0.184(0.027) \\
    Molecule only & 0.319(0.106) & 0.355(0.028) & 0.637(0.069) & 0.188(0.029) \\
    Hard + Soft $\lambda$=0.5 & 0.356(0.046) & 0.342(0.017) & 0.624(0.039) & 0.178(0.013) \\
    Only Soft $\lambda$=2.0 & 0.360(0.072) & 0.331(0.019) & 0.603(0.062) & 0.177(0.020) \\
    No Orthogonal & 0.360(0.069) & 0.347(0.014) & 0.617(0.056) & 0.177(0.019) \\
    Hard + Soft $\lambda$=1.0 & 0.387(0.094) & 0.331(0.021) & 0.632(0.069) & 0.169(0.026) \\
    Inter & 0.385(0.067) & \textbf{0.328(0.017)} & 0.638(0.055) & 0.170(0.019) \\
    NOSE & \textbf{0.389(0.052)} & 0.333(0.021) & \textbf{0.657(0.029)} & \textbf{0.169(0.014)} \\
    \bottomrule
  \end{tabular}
  \caption{Ablation study: mixture prediction on intensity.}
  \label{tab:ablation_10}
\end{table*}

\begin{table*}[t]
  \centering
  \small
  \begin{tabular}{lcccc}
    \toprule
    & \multicolumn{4}{c}{\textbf{Mixture Pleasantness}} \\
    \cmidrule(lr){2-5}
    \textbf{Method} & $R^2$ $\uparrow$ & MAE $\downarrow$ & Pearson $r$ $\uparrow$ & MSE $\downarrow$ \\
    \midrule
    Adapter (Desc: 10.00M, Rec: 10.00M) & 0.404(0.260) & 0.671(0.133) & 0.781(0.068) & 0.681(0.298) \\
    Hard + Soft $\lambda$=0.5 & 0.509(0.053) & 0.610(0.042) & 0.753(0.021) & 0.561(0.060) \\
    Molecule only & 0.509(0.047) & 0.614(0.023) & 0.795(0.035) & 0.561(0.054) \\
    Adapter (Desc: 29.17M, Rec: 10.00M) & 0.521(0.127) & 0.591(0.066) & 0.790(0.032) & 0.547(0.145) \\
    Hard + Soft $\lambda$=0.1 & 0.529(0.151) & 0.582(0.091) & 0.786(0.028) & 0.538(0.172) \\
    Only Soft $\lambda$=2.0 & 0.549(0.018) & 0.603(0.013) & 0.792(0.021) & 0.516(0.021) \\
    Inter & 0.558(0.108) & 0.586(0.061) & 0.788(0.063) & 0.505(0.123) \\
    Only Hard & 0.547(0.083) & 0.582(0.029) & 0.806(0.040) & 0.518(0.094) \\
    Description only & 0.555(0.050) & 0.574(0.028) & 0.798(0.020) & 0.508(0.058) \\
    Hard + Soft $\lambda$=1.0 & 0.574(0.087) & 0.572(0.044) & 0.777(0.054) & 0.487(0.099) \\
    Inter + Weak & 0.566(0.106) & 0.565(0.075) & 0.788(0.062) & 0.496(0.121) \\
    No Orthogonal & 0.562(0.088) & 0.577(0.067) & 0.807(0.029) & 0.501(0.101) \\
    Inter + Intra & 0.606(0.036) & 0.557(0.021) & 0.821(0.030) & 0.450(0.041) \\
    Receptor only & \textbf{0.648(0.047)} & 0.536(0.026) & 0.836(0.039) & \textbf{0.402(0.054)} \\
    NOSE & 0.636(0.047) & \textbf{0.534(0.033)} & \textbf{0.846(0.012)} & 0.416(0.054) \\
    \bottomrule
  \end{tabular}
  \caption{Ablation study: mixture prediction on pleasantness.}
  \label{tab:ablation_11}
\end{table*}

\end{document}